\documentclass[11pt]{article}

\usepackage{natbib}
\usepackage[margin=2.5cm]{geometry}
\usepackage{comment}
\usepackage{tipa}
\usepackage{graphicx}
\usepackage{setspace}
\usepackage[none]{hyphenat}
\usepackage{enumerate}
\usepackage[hidelinks]{hyperref}

\author{Patrycja Strycharczuk, Sam Kirkham, Emily Gorman \& Takayuki Nagamine}
\title{Towards a dynamical model of English vowels.\\
Evidence from diphthongisation}
\date{}

\begin{document}

\begin{centering}
Towards a dynamical model of English vowels. Evidence from diphthongisation\\
\vspace{1cm}
Patrycja Strycharczuk$^{1}$, Sam Kirkham$^2$, Emily Gorman$^2$ and Takayuki Nagamine$^2$

\vspace{2cm}
\end{centering}

\noindent
$^1$Linguistics and English Language\\
University of Manchester\\
United Kingdom\\

\noindent
$^2$Linguistics and English Language\\
Lancaster University\\
United Kingdom\

\vspace{1cm}
\noindent
Corresponding author's contact details:\\
Department of Linguistics and English Language\\
University of Manchester\\
Oxford Rd\\
Manchester\\
M13 9PL
United Kingdom\\AH
patrycja.strycharczuk@manchester.ac.uk

\abstract{Diphthong vowels exhibit a degree of inherent dynamic change, the extent of which can vary synchronically and diachronically, such that diphthong vowels can become monophthongs and \emph{vice versa}. Modelling this type of change requires defining diphthongs in opposition to monophthongs. However, formulating an explicit definition has proven elusive in acoustics and articulation, as diphthongisation is often gradient in these domains. In this study, we consider whether diphthong vowels form a coherent phonetic category from the articulatory point of view. We present articulometry and acoustic data from six speakers of Northern Anglo-English producing a full set of phonologically long vowels. We analyse several measures of diphthongisation, all of which suggest that diphthongs are not categorically distinct from long monophthongs. We account for this observation with an Articulatory Phonology/Task Dynamic model in which diphthongs and long monophthongs have a common gestural representation, comprising two articulatory targets in each case, but they differ according to gestural constriction and location of the component gestures. We argue that a two-target representation for all long vowels is independently supported by phonological weight, as well as by the nature of historical diphthongisation and present-day dynamic vowel variation in British English. \\

}

\doublespacing

\newpage
\section{Introduction} \label{intro}
Diphthong vowels are characterised by some degree of inherent dynamic change over the course of the vowel, a salient property that needs to be captured by models of vowel production. The traditional theoretical device used to reflect this phonetic property of diphthongs on a more abstract level is compositionality. By `compositionality' we mean modelling diphthongs as consisting of two component elements, with no further assumptions about the nature of these elements. Note that this is a broader sense than adopted by some other works, such as \cite{hsieh2017}, who defines `compositionality' as being composed from two elements that function independently within the same system. Compositionality is evoked by the etymology of the word \emph{diphthong} (from Greek d\'{i}phthongos, `two sounds'), it has been incorporated into phonological models of diphthongs across different theoretical frameworks, it is implicit in IPA transcription, and it is also implicit in the common descriptive practice of using two time points to represent diphthongs. Crucially, the compositional nature of diphthongs is typically conceived of in opposition to monophthongs, which consist of a single component element, represented by a single IPA symbol, and commonly reduced to a single time point in phonetic measurements, normally the acoustic midpoint. While the descriptive tradition may be a matter of convenience, theoretical models that employ a distinction between compositional diphthongs and single component monophthongs predict categorical differences between the two types of sounds. 

\subsection{Compositionality in models of diphthongs} \label{comp}

An example of a phonological model positing a distinct representation for monophthongs and diphthongs is the Autosegmental Phonology model by \cite{goldsmith1990}, who proposes that diphthongs consist of two segments (two root nodes), each linked to a single timing unit, whereas monophthongs consist of a single segment (a single root node), which however, may be linked to two timing units, as is the case with long monophthongs. This constitutes a systematic structural difference between diphthongs and long monopthongs, in that either two or one root nodes are present. 

A similar distinction is made by some Articulatory Phonology / Task Dynamics (AP/TD) models of vowel representation, except the relevant component element are not segments, but gestures. Within AP/TD, articulatory representations consist of discrete gestures \citep{browman1986, browman1992, saltzman1989}. Each gesture has a specific spatio-temporal target, whereas transitions arise from movement from one target to another, constrained by a degree of overlap between the gestures. For diphthongs specifically, multiple accounts postulate two component elements. For example, \cite{marin2007} proposes that complex nuclei effects in diphthongs, such as specific patterns of syllable weight and stress alternations, can be accounted for in AP/TD through a mechanism of gestural coupling originally proposed to explain syllable organisation. She develops a model for Romanian mid diphthongs /ea/ or /oa/, in which the diphthongs are composed of two distinct synchronous (in-phase) vocalic gestures. Variation in the degree of diphthongisation is modelled using blending strength. If the two gestures have an equal blending strength, the output is a monophthong, intermediate between the two gestures. However, increasing the blending strength of one of the two gestures produces a diphthong characterised by observable inherent change. Marin proposes that variation in blending strength arises through additional factors, such as stress or speech rate, and this accounts for the phonetic variation observed in Romanian. This model is not claimed to be a universal representation of all diphthong vowels, but it is suggested that it could be extended to some types of diphthongs in other languages. One of the examples provided is the ongliding sequence /\textipa{ju:}/ in American English (as in \emph{beauty} or \emph{few}), which is analysed as a sequence of two vocalic elements that are synchronously coordinated. In contrast, offgliding diphthongs (like \textsc{price} or \textsc{choice}) are said to consist of a nucleus vowel and a coda glide that are coupled sequentially to each other (anti-phase). A key argument for distinguishing between the two types of diphthongs is syllable weight. The onglide /j/ in /\textipa{ju:}/ is non-moraic, which supports the in-phase organisation between the two elements. In contrast, offgliding diphthongs are bimoraic, which supports a coda-like sequential organisation.\footnote{There is a further complicating factor in that /j/ and /w/ in offgliding diphthongs are considered consonantal in this account. This assumption is somewhat controversial, as \cite{burgdorf-tilsen2021} show that difference between vowels and glides can be modelled through their temporal properties rather than an inherent spatial distinction. The issue of whether glides are consonantal or not is not crucial to the idea that ongliding diphthongs differ from offgliding diphthongs in their gestural organisation.} Marin does not specifically discuss the representation of long monophthongs vis-\`{a}-vis diphthongs, but she models the offglide /u\textipa{:}/ in \emph{few}, which is long, as corresponding to a single gesture. This suggests that the primary structural difference between diphthongs and long monophthongs is the presence of two vs. one gesture, a gestural equivalent of the autosegmental distinction discussed above.

This type of distinction is also adopted in TADA (Task Dynamics Application), the computational model of AP/TD \citep{nam-etal2004}. In its current implementation, TADA represents diphthongs, such as \textsc{price} using two targets: a nucleus and an offglide; whereas  monophthongs, such as \textsc{thought}, are represented using a single gestural target. Notably,  high long vowels (\textsc{fleece, goose}) are represented using two gestures, similarly to canonical diphthongs. We note that this representation may not represent a firm theoretical commitment, and we can only speculate about the rationale behind adopting it, but it is likely that two gestural targets are used for \textsc{fleece} and \textsc{goose} to accommodate the small degree of dynamic change that characterises such vowels, i.e. a form of gradient diphthongisation. 

\subsection{Gradience in diphthongisation} \label{gradient}
Diphthongisation frequently shows gradient characteristics. Below, we discuss some manifestations of this phenomenon, before considering the relevance of gradient diphtongisation to diphthong compositionality. 

Firstly, it is very easy to find examples of vowels that are intermediate between monophthongs and diphthongs, i.e. vowels that show a perceptible degree of dynamic change, which is however small. For example, \cite[p.11]{sweet1910} makes a distinction between `half-diphthongal' vowels sounds [ei] (\textsc{face}) and [ou] \textsc{goat}, as opposed to `fully-diphthongal' [ai] (\textsc{price}) and [oi] (\textsc{choice}). A three-way split according to degree of diphthongisation is also proposed by \cite{lehiste1961} for American English, who use a systematic acoustic approach to quantify degree of diphthongisation. They distinguish between diphthongs, which have two acoustic targets (defined as two acoustic steady states), monophthongs, which have one, and an intermediate category of glides. Glides are characterised by a single target, but also by formant transitions that are inherent to the vowel (as opposed to being transitions from the preceding or into the following consonants). The vowels identified as glides in that way are American English \textsc{face}, \textsc{goat} and \textsc{nurse}. In contrast,  \textsc{price}, \textsc{mouth} and \textsc{choice} are classified by \citeauthor{lehiste1961} as diphthongs. A number of sources adopt a similar three-way distinction that acknowledges a special status of  \textsc{price}, \textsc{mouth} and \textsc{choice}  in English as `true', `full' or `phonemic' diphthongs \citep{assmann1982, jacewicz2012, morrison2007}. English is not unique in having varying levels of diphthongisation in its vowel inventory. For example, in Dutch, a distinction has been proposed between `potential diphthongs', /ei/, /\o y/ and /ou/, and `essential' diphthongs,  /\textipa{EI}/, /\oe y/ and /\textipa{\textscripta u}/ \citep{collins2003}.

Since gradient diphthongisation is by definition phonetically small, discrepancies arise between sources, according to which vowels are treated as diphthongs when a binary monophthong-diphthong split is adopted. Consider, for example, the transcription of the \textsc{fleece} and \textsc{face} vowels in three varieties of English, as captured by the IPA Illustrations of Received Pronunciaction (RP; \citealt{roach2004}), American English \citep{ladefoged1999} and Australian English \citep{cox2007}. The \textsc{fleece} vowel is transcribed as monophthongal /\textipa{i(:)}/ in all three, although \citealt{ladefoged1999} suggests that /\textipa{i\textsuperscript{j}}/ is a possible transcription. In comparison, \textsc{face} is transcribed as /\textipa{eI}/ in \cite{roach2004}, as /\textipa{e}/ in \cite{ladefoged1999}, although /\textipa{e\textsuperscript{j}}/ is also mentioned as a possibility, and as /\textipa{\ae I}/ in \cite{cox2007}. It is somewhat unclear to what extent these discrepancies arise due to differences in transcription conventions, and to what extent they are driven by differences in phonetic quality of these vowels between different varieties. Some diphthongisation of \textsc{fleece} is present in all three varieties, and it is quite salient in Australian English \citep{cox2023}, yet the transcription tends not to reflect it. Thus, transcribing \textsc{fleece} as a monophthong seems to be driven by convention, as indeed acknowledged by \cite{ladefoged1999}. In contrast, \textsc{face} is transcribed variably with one or two symbols, and  here some phonetic differences are likely at play. For example, \textsc{face} is clearly more diphthongised in Australian English compared to American English, so this may explain why the vowel would be classified as a diphthong in the former case and as a monophthong in the latter. 

Variable diphthongisation is a perennial feature of English, such that the same vowel phoneme may have a monophthongal quality in one variety, but diphthongal in another, and we also find that monophthongs and diphthongs can morph into each other fairly freely. Northern Anglo-English, which we shall focus on in this study, has been noted for a relatively large degree of variation in the degree of diphthongisation. The monophthongal quality of \textsc{face} and \textsc{goat} is mentioned as a distinguishing accent feature in descriptions of traditional varieties of Lancashire and Yorkshire English \citep{wells1982.2}, and also a possible variant in Newcastle English \citep{watt2002}. A monophthongal quality of \textsc{price} can also be a feature in these dialects, as well as in Liverpool English \citep{watson2007}. In contrast, \textsc{face}, \textsc{price} and \textsc{goat} are considered diphthongs in Manchester English \citep{baranowski2015me}. Alongside the local varieties, many speakers in the North of England speak a pan-regional variant, General Northern English, in which \textsc{price}, \textsc{mouth}, \textsc{choice}, \textsc{face} and \textsc{goat} are acoustically clearly diphthongal \citep{strycharczuk2020}. According to \cite{honeybone2007}, \textsc{square} is generally a monophthong in the North of England, as confirmed by \cite{strycharczuk2020}, who also report a monophthongal quality of \textsc{near} in Manchester English, but not in other urban locations in the North of England. \cite{williams2014} note a diphthongal quality of \textsc{goose} in a sample of speakers from Sheffield. While these reports are based on different criteria and different methodologies, they clearly suggest considerable dynamic variability, as some vowels are likely to cross the divide between monophthongs and diphthongs, depending on the dialect, speaker, and potentially style.

\subsection{Gradient diphthongisation as a challenge to diphthong compositionality} \label{theoretical_issues}

So far we have demonstrated that gradient diphthongisation is common in English, which makes it difficult to separate monophthongs from diphthongs. This has practical consequences for choices guiding phonetic analysis, and theoretical consequences for phonological models of vowels.

We have already discussed the issues related to transcription: gradient diphthongisation forces the transcriber to make a somewhat arbitrary choice between using one or two symbols. Phonetic transcription is, of course, to some extent arbitrary, but when it comes to vowel dynamics, it may have analytical consequences. It is common methodological practice to represent vowel quality in diphthongs using two time points, whereas a single time point is used for monophthongs. The IPA illustrations discussed above in Section \ref{gradient} serve as an example.  It is also common to follow earlier description and transcription conventions in deciding which vowels are monophthongs and which vowels are diphthongs. In cases where gradiently diphthongised vowels are treated as monophthongs, this results in a loss of important dynamic information. The problem of dynamic reduction has received considerable attention in recent years, with many phonetic studies of vowel reduction moving away from temporal reduction, and not attempting an \emph{a priori} distinction between monophthongs and diphthongs. For example, many recent studies analyse entire formant trajectories, using techniques such as Generalised Additive Modelling, Discrete Cosine Transformation, or Smoothing Splines ANOVA (see \citealt{cox2023} for a recent overview).

While methodological practice can simply avoid dealing with gradient diphthongisation by treating monophthongs and diphthongs alike, representational models of vowels must take a stance on how to distinguish monophthongs from diphthongs, whilst also capturing gradient diphthongisation. If we assume that the structural difference between monophthongs and diphthongs lies in the underlying  number of gestures, then diachronic diphthongisation would involve insertion of a gestural target, whereas monophthongisation would involve deletion of a target.\footnote{The generalisation and the problem are much the same for an autosegmental representation which uses the number of root nodes to distinguish between monophthongs and diphthongs.} Either insertion or deletion would constitute major restructuring, and from that point of view, we might expect such changes to be somewhat constrained. However, as we have already discussed, the opposite is the case: such changes are very common, and it is also common to find stylistic variation within a single speaker between a monophthongal and diphthongal version of the same vowel phoneme. Furthermore, changes in the degree of diphthongisation usually do not entail changes in phonological patterning, such as phonological weight, or phonotactic constraints. In English, long monophthongs and diphthongs function structurally in the same way. Thus, from the point of view of phonology, as well as from the point of view of variation and change, there is no strong argument to distinguish between long monophthongs and diphthongs.

Some existing phonological accounts include gradiently diphthongised vowels in the diphthong category, such that effectively, most long vowels are treated as diphthongised. \cite{popescu2022} propose that diphthongised long vowels comprise two targets, similarly to diphthongs. They also propose that different degrees of diphthongisation may emerge, depending on how similar the component targets are. When the component targets are different, the vowel is clearly diphthongal. In contrast, when the component targets have a similar constriction location, the result may be a slightly diphthongised vowel. This proposal extends a two-target representation to instances of gradient diphthongisation, similarly to the TADA approach (Section \ref{comp} above). 

Some variationist literature goes one step further. \cite{labov2006} adopt a two-way split within the North American English vowel inventory: short vs. long vowels. All long vowels are treated as inherently diphthongal, and it is said that a degree of diphthongisation may emerge for all long vowels in the final position. The diphthongal interpretation of long vowels is also reflected in those vowels being represented by a pair of symbols, whereas a single grapheme is used for short vowels. This system is said to be closely based on \cite{wells1982.3}, who proposes a split between long and short vowels, with  further subdivision of long vowels into upgliding and ingliding ones. While this distinction is not reflected in the Wells's transcription system for American English, this notational decision is attributed by \cite{labov2006} to convention rather than to mark absence of any diphthongisation. Interestingly, seeds of the idea that diphthongs and long monophthongs are structurally the same may be found even further back. \citet[p.173]{trubetzkoy1971} discusses a case of Slovak vowels, for which he argues that``long nuclei are interpreted as monosyllabic combinations of two like vowels''.

An alternative theoretical proposal for dealing with gradient diphthongisation could be enriching the representation with dynamic detail, such that the phonology stores continuous dynamic information rather than abstract targets from which the dynamics emerge in speech. This type of rich representation would not need to make a distinction between diphthongs and monophthongs, instead treating all vowels as dynamic.  \cite{xu2023} present a model along these lines, specifying diphthong representations through a combination of targets and slopes that determine the direction and range of articulatory movement. This type of articulatory modelling represents a radical departure from diphthong compositionality. While the model has not yet been fully developed or validated, it represents a possible conceptual direction for theories of vowel representation that seems compatible with gradient diphthongisation.

The debate concerning diphthong compositionality in articulation closely resembles a similar debate about the status of diphthongs in the perceptual domain. The observations concerning the compositional nature of diphthongs were originally based on auditory properties of vowels, and auditory judgements are still frequently used when classifying vowels as diphthongs or monophthongs. Perceptual research confirms that diphthong vowels can be successfully identified when reduced to two time points, but not one, whereas increasing temporal resolution further does not substantially improve vowel identification \citep{jibson2022}. While these findings are apparently in line with the interpretation that diphthongs have two distinct perceptual targets, this advantage of two point models over single point ones may not be not unique to diphthongs: a similar advantage is observed for ostensible monophthongs by \cite{hillenbrand2013} and \cite{jibson2022}, although \cite{harrington1994} present divergent findings. 

In summary, the idea that diphthongs have two component elements is very well established in phonology and phonetics, as evident from theoretical models and common methodological practice in phonetics. However, any model of vowel representation that incorporates this idea must specify which vowels are diphthongs, and should therefore be modelled as containing two elements. Such a decision may not be straightforward for any given vowel inventory, because diphthongisation appears to be phonetically gradient. Gradient diphthongisation represents a potential area of overlap between monophthongs and diphthongs in which a vowel shows a small degree of inherent change. Based on the proposals in previous literature, there are broadly three ways of capturing gradient diphthongisation in models of vowel representation. 
\begin{enumerate}[i.]
    \item Canonical diphthongs are modelled as compositional, as are instances of gradient diphthongisation. In contrast, canonical long monophthongs are modelled as having one component element.  This approach incorporates a structural distinction between long monophthongs and diphthongs, and it is consistent with \cite{popescu2022} and the current TADA model for American English.
    \item All long vowels are modelled as compositional, i.e. all long vowels are inherently diphthongs. This possibility presupposes no categorical distinction between monophthongs and diphthongs, and it is consistent with \cite{labov2006} for American English.
    \item All vowels are specified for target and trajectory of movement. This approach requires no separation between monophthongs and diphthongs, and it also entails that diphthongs are not inherently compositional.
\end{enumerate}

Our study sets out to inform the discussion about the empirical accuracy of these approaches and their theoretical advantages and disadvantages. The novel perspective we offer comes from a systematic articulatory study of vowel diphthongisation and from articulatory modelling.

\subsection{This study}

We present a systematic articulatory investigation of diphthongisation across the long vowel subsystem, based on electromagnetic articulography (EMA) data from six speakers of Northern Anglo-English. This variety was chosen because it is rich in potential instances of gradient diphthongisation, as reflected by the considerable dynamic variation. We analyse the dynamic properties of all long vowels in order to quantify their relative degree of diphthongisation, using both articulatory and acoustic metrics. The two main research questions guiding this analysis are as follows. 

\begin{enumerate}
\item Are articulatory properties of diphthongs consistent with a model in which diphthongs comprise two distinct gestural targets?
\item Are diphthongs categorically distinct from monophthongs, or is diphthongisation is gradient in the articulatory domain? We conceive of gradience as phonetic continuity in measures that capture degree of diphthongisation. Thus, we operationalise our question as a classification problem: Can we systematically distinguish monophthongs from diphthongs, using articulatory and acoustic diagnostics?
\end{enumerate}

As far as we know, no study to date has documented articulatory properties of diphthongisation across multiple vowels in the same language/variety. This creates a major gap in our understanding of the nature of diphthongisation, because gradient diphthongisation may in principle arise in acoustics and perception while there is a categorical difference between diphthongs and monophthongs at the articulatory level. A possible conceptualisation of this dichotomy is offered by \cite{strange1989}, who argues that listeners are sensitive to dynamic information that emerges from gestural vowel dynamics, which includes the gestural target, but also the opening and closing phase. In this sense, vowel targets could be seen as dynamic from the perceptual point of view, but categorical at the underlying articulatory level, manifested as the presence or absence of component articulatory gestures. 

An additional argument for considering articulatory evidence is the fact that the presence or absence of articulatory targets has a stable empirical correlate, under the core assumptions of Articulatory Phonology. In Articulatory Phonology, gestures are abstract units of organisation, which are however, measurable because they have systematic physical correlates. The key correlate  of a gesture is movement towards a specific articulatory position (gestural maximum), followed by a change of direction of the articulatory movement (gestural release). This framework also provides a principled way of reconciling categorical and gradient aspects of speech, as it explicitly models continuous speech signal as emerging systematically from a combination of gestural targets which are categorical units. In this context, we ask whether we can reconstruct the underlying number of gestural targets from the articulatory data, and whether this allows us to classify all vowels as having one or two component targets. This question informs our theoretical perspective on diphthong compositionality, understood as diphthongs being composed of two gestural targets (two independently timed articulatory gestures). We make the distinction between a target and a gesture, because the same sound can be composed of multiple simultaneous gestures, such as the movement of the tongue dorsum and the lips in case of back rounded vowels. While there are two gestures in this case, they overlap closely and can thus be presumed to contribute towards a single target. 

Importantly, gestural targets may be present underlyingly but they may not be identifiable in the resulting articulatory signal due to factors such as gestural overlap (one gesture masking another gesture), or gestural undershoot (gesture being reduced). Thus, we supplement our articulatory data with computational modelling manipulating the values of articulatory parameters, and we evaluate various models against the empirical data obtained in the experiment.

\section{Materials and method} \label{method}
\subsection{Stimuli}
The stimuli recorded in the experiment included a full set of English long vowels in an open syllable, preceded by the voiced bilabial stop /b/. The specific words were: \emph{bay, buy, boy, bough, beau, beer, bear, bee, burr, bar, bore, boo}. Note that present-day Northern Anglo-English is generally non-rhotic, with the exception of some areas in Lancashire \citep{turton2023}. None of our participants pronounced coda rhotics. We used an initial bilabial, because it provides a relatively neutral context for all the vowels, as far as the tongue movement is concerned, and also because all combinations of long vowels with a preceding /b/ correspond to real words in the English lexicon. We used CV words, because in the absence of a following coda, we can be certain that any offglide movement we observe is inherent to the vowel, and not due to coarticulation. Since only phonologically heavy vowels can occur in this context, we did not include short monophthongs. The target words were embedded in the carrier phrase: \emph{She says X}. 
\subsection{Participants}

Six female speakers aged between 19--21 years old ($\overline{x}$ = 19.17, $\sigma$ = 0.98) took part in the experiment. All participants reported normal speech and hearing, and all were monolingual L1 speakers of English. All speakers were born and grew up in the north of England. Specifically, five speakers lived in the region spanning Lancashire and Greater Manchester from birth until the time of the experiment, while one speaker lived in Sheffield until moving to Lancaster at the age of 18. All speakers had an unambiguously northern English phonological system and all used a notably higher proportion of regionally-marked features than is typical of General Northern English as described in \cite{strycharczuk2020}. Each participant was reimbursed \pounds30 for taking part in the study, which lasted around 1.5 hours in total.

\subsection{Procedure}

Electromagnetic articulography (EMA) data were acquired using a 16-channel Carstens AG501 system, recording at a sampling rate of 1250 Hz. Sensors were attached to the tongue at 1cm behind the tongue tip (TT), as far back as possible on the tongue dorsum (TD), and an additional sensor located equidistant between the TT and TD sensors. Sensors were also attached to the vermilion border of the upper (UL) and lower (LL) lips, as well as the lower gumline. Reference sensors were attached to the gumline of the upper incisors, bridge of the nose, and on the right and left mastoids behind the ears. All sensors were attached midsagittally, except for the sensors behind the ears. Simultaneous ultrasound tongue imaging data were collected alongside the audio and EMA data, with an additional three sensors attached to the ultrasound probe, but the ultrasound data are not analysed in this study (see \citealt{kirkham2023_co} for more detail on the co-registration set-up). We recorded the location and orientation of the occlusal plane for each speaker by asking them to bite down onto a bite plate, which also had three EMA sensors attached to it. The audio signal was recorded using a Beyerdynamic Opus 55 microphone attached to a plastic ultrasound probe stabilisation headset, which was being used for simultaneous ultrasound data collection. The microphone signal was pre-amplified using a Grace Design m101 pre-amplifier and digitised at 48 kHz with 16-bit quantisation.

Most speakers produced four repetitions of the stimuli, except f05, who produced five. At the processing stage, it became apparent that the Tongue Dorsum displacement data were considerably out of range for multiple blocks produced by speakers f03 and f06. This suggests that the sensor became detached without the participant or ourselves noticing. We discarded these data from the articulatory analysis, which left a single usable set of repetitions for f03, and two repetitions for f06. The total number of tokens used in the articulatory analysis was 242. We included all the available data (306 tokens) in the acoustic analysis.

\subsection{Data processing}

EMA recordings were downsampled to 250 Hz and position calculation was carried out using the Carstens default algorithms. Head correction and bite plate rotation were applied to each data sample, with head correction for each speaker optimised based on the best combination of reference sensors that reduced the RMS error across the entire session. Reference sensors were filtered using a 5 Hz low-pass Kaiser-windowed filter, and articulator sensors were filtered using a low-pass Kaiser-windowed filter with 40 Hz pass and 50 Hz stopband edges. We further filtered articulator sensors for the specific analyses reported below in Section \ref{analysis}.

\subsection{Analysis} \label{analysis}
The audio data were forced-aligned using the Montreal Forced Aligner \citep{mcauliffe17}. The boundaries were then manually corrected in Praat \citep{praat6214} by a research assistant, with specific attention paid to the acoustic boundaries of the vowel. The vowel onset was placed at the end of the burst for the preceding plosive, which typically coincided with the onset of a visible formant structure and the offset of voicing. The end of the vowel was marked at the offset of voicing. 

The segmentation was used as the basis for articulatory and acoustic analysis. For the articulatory analysis, we focus on the movement of two key sensors: Tongue Dorsum (TD) and Upper Lip (UL). Displacement of the Tongue Dorsum allows for a systematic parsing of vowel gestures, as previously shown by \cite{blackwood2017} and \cite{sotiropoulou2020}. In addition, some vowels are crucially modified by the movement of the lips. In our analysis of the lip movement, we focus on lip protrusion as a correlate of rounding. Lip protrusion can be defined as the horizontal displacement of the lips, although the exact definition varies across the literature \citep{georgeton2014}. We focus on the displacement of the upper lip, because we find that the horizontal displacement of the lower lip was systematically affected by the jaw movement, whereas the movement of the upper lip appears more independent. Here, we wanted to use displacement vectors that are as orthogonal as possible to avoid analytical artefacts that could arise from different displacement vectors capturing the same movement, as we combine them into a joint measure. 
The sensor displacement data were $z$-scored within speaker for normalisation.

The sensor displacement data were extracted for the portion corresponding to the acoustic duration of the vowel, followed by a fixed 75ms window at the end. We included this window because we find that movement of the tongue and the lips typically continues beyond the offset of voicing at the end of a phrase. Much of our analysis focuses on the first derivative of the sensor displacement data, i.e. velocity. In order to calculate the velocities, we further smoothed the scaled displacement values using a low-pass Butterworth filter (cutoff frequency $=$ 10 Hz). Tangential  velocity was then derived, combining the horizontal-vertical Tongue Dorsum sensor displacement, and the horizontal displacement of the Upper Lip.

Acoustic data were analysed in Praat. We extracted formant trajectory values for each vowel, using Fast Track \citep{barreda2021}. The settings we used were: lowest analysis frequency $=$ 5kHz, highest analysis frequency $=$ 7 kHz, Number of steps = 20, Coefficients for formant = 5, Number of formants = 3, Number of bins = 5, Statistic = median. The formant measurements were sampled at every 2ms (equivalent to 500 Hz). Similar as in the articulatory analysis, we also analysed the first derivative of the formant change. We $z$-scored the formant measurements within speaker, and smoothed the formant trajectories using a low-pass Butterworth filter (cutoff frequency $=$ 10 Hz). 

Since our study is concerned with quantifying degree of diphthongisation, it is important to comment on how such a degree can be measured. Several acoustic studies do this by measuring the degree of overall formant displacement. The specific measure that can be used is the Euclidean distance between F1 and F2 values at the onset and offset of the vowel, or an interval equivalent to the 80\% portion of the vowel \citep{fox2009, haddican2013, reed2014}. This measure can also be refined to account for more complex trajectory shape as a sum of Euclidean distances sampled from a number of windows within the vowel \citep{fox2009}. We use the simpler version, relying on two time points, and we also extend the same approach to articulatory data. We calculate the Euclidean distance between two articulatory positions at pre-defined time points in order to capture the overall degree of articulatory displacement within a vowel. Section \ref{art_dip} below provides more detail on how the articulatory Euclidean distance was calculated, and Section \ref{form_dis} does the same for the acoustic Euclidean distance.

Euclidean distances can capture the overall degree of acoustic and articulatory change, and they are thus inherently well-suited to capturing gradient aspects of diphthongisation: canonical diphthongs are characterised by more inherent change, compared to canonical monophthongs, with gradient diphthongisation creating in-between patterns. However, since our main research question is whether we can classify all vowels within a system as having one or two component gestural targets, we also need a procedure for identifying the number of gestures and an associated measure.

In Articulatory Phonology, gestural targets are typically identified using the first derivative of displacement data, i.e. velocity. This is based on the observation that the rate of change in movement towards a target is associated with a change of direction in articulatory displacement, which corresponds to a local minimum in the associated tangential velocity profile. The approach underlies gestural parsing, as implemented in MVIEW, widely used software for analysing gestural properties using EMA data \citep{tiede2010}. We  analyse velocity trajectories to establish whether we find evidence of two component targets for diphthong vowels, and whether we can classify all vowels as having one or two discernible component gestures.  Section \ref{art_vel} provides a more detailed explanation of the patterns we find.

We also extend the same approach to analysing the rate of change in formant trajectories. Unlike in articulatory studies, the first derivative is not a common measure in acoustics, so we do this largely on an exploratory basis, and to keep our articulatory and acoustic analysis as comparable as possible. However, there are several arguments for analysing formant velocity in addition to formant displacement (which we captured using Euclidean distance). Velocity is inherently a measure of rate of change, and as such, it can straightforwardly reflect the presence of a formant steady state (an interval of low velocity) vs. change in formant trajectories (a sustained rise in velocity). Furthermore, the measure captures the degree of change irrespective of the direction of movement, and it allows us to combine information from multiple formants into a single value. In order to obtain this measure, we calculated the rate of change per unit time for the smoothed F1 and F2 trajectories. Tangential F1-F2 velocity was calculated based on these measures, defined as the square root of the sum of squared F1 and F2 velocities. Section \ref{form_vel} illustrates the resulting patterns, and suggests that tangential velocity is well-suited to quantifying the global rate of change in acoustic dynamics.

\subsection{Data availability statement}

All the data and code presented in this paper are available as an Open Science Framework repository at \url{https://osf.io/gub32/}. The repository also contains the simulations presented in Section \ref{model}.

\section{Results} \label{results}
\subsection{Articulatory displacement} \label{art_dip}
\begin{figure}[htbp]
\begin{center}
\includegraphics[scale=1]{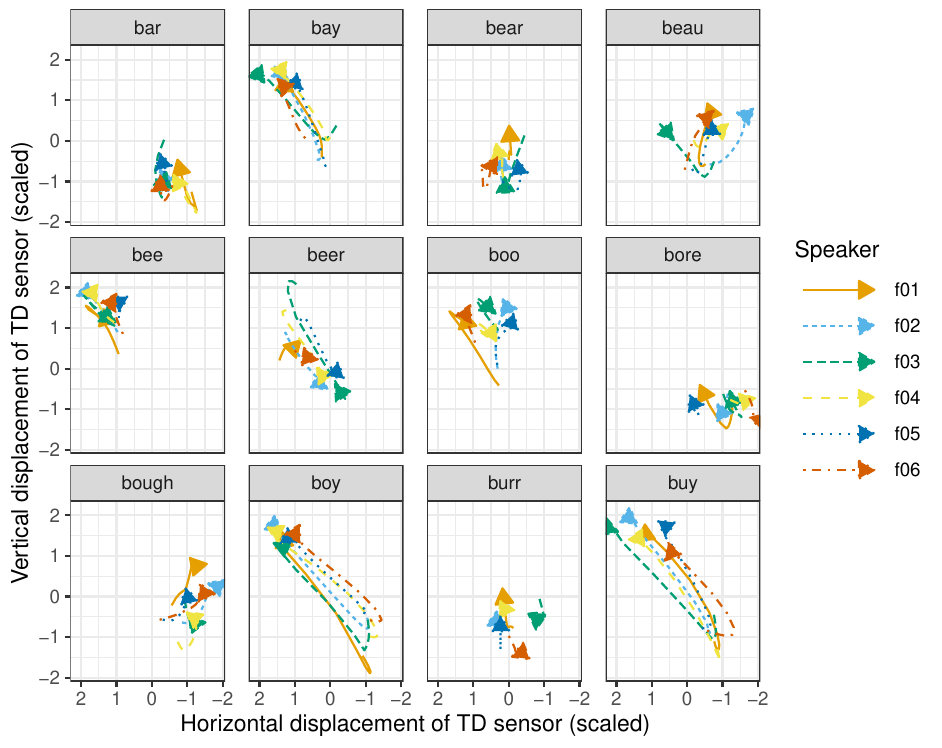}
\caption{Mean by-speaker trajectory of TD sensor displacement for each vowel. The beginning and the end of each arrow correspond to the acoustic onset and offset of the vowel respectively.}
\label{fig:td_disp}
\end{center}
\end{figure}

We begin the presentation of results with a visual overview of articulatory displacement. Figure \ref{fig:td_disp} shows by-speaker mean TD displacement values for each item.  The trajectories correspond to the acoustic duration of the vowel. The means were obtained using Generalised Additive Modelling (GAM). As expected, we see the greatest degree of TD displacement for vowels in \emph{boy} and \emph{buy}, followed by \emph{bay}, \emph{beau} and \emph{beer}. Canonical monophthongs like \emph{bar} or \emph{burr} show overall least TD displacement, but there is some TD movement associated with these vowels. There is relatively limited TD displacement in \emph{bough}, despite this vowel's robust acoustic and perceptual diphthongisation. This is likely due to the fact that the perceived change in this vowel is strongly affected by lip rounding, whereas the associated lingual movement is limited.

\begin{figure}[htbp]
\begin{center}
\includegraphics[scale=1]{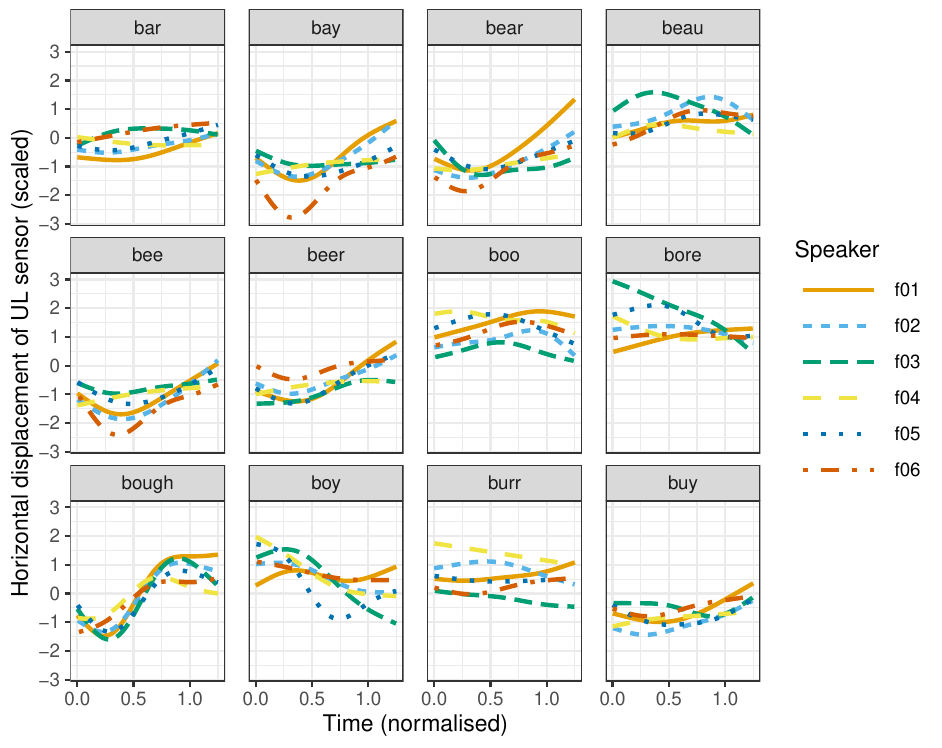}
\caption{Mean by-speaker trajectory of UL sensor displacement for each vowel, relative to normalised time}
\label{fig:ul_disp}
\end{center}
\end{figure}

The role of the lips in diphthongisation is confirmed when we consider the horizontal displacement of the upper lip in normalised time, plotted in Figure \ref{fig:ul_disp}. The figure shows by-speaker GAM smoothed mean trajectory of horizontal Upper Lip displacement in normalised time. Overall, the displacement values are higher for rounded vowels like \emph{boo} and \emph{bore}, compared to unrounded vowels like \emph{bee}. We can also see that some vowels are characterised by inherent change in lip displacement. This is especially prominent for \emph{bough}, where we can see a forward movement of the upper lip, consistent with a rounding gesture, and in \emph{boy}, where we find the opposite: the UL sensor moves backwards through the vowel articulation, which can be interpreted as an unrounding gesture.

\begin{figure}[htbp]
\begin{center}
\includegraphics[scale=1]{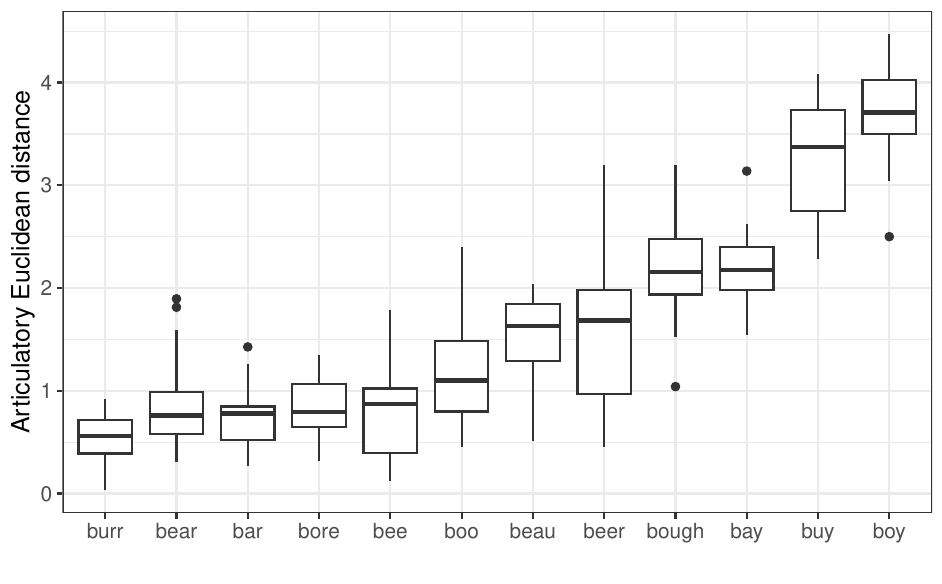}
\caption{The articulatory Euclidean distance depending on the vowel}
\label{fig:art_ED}
\end{center}
\end{figure}

In order to systematise these observations and to account for the dorsal and labial displacement using a single measure, we calculated the Euclidean distance between 10\% and 90\% of the vowel in a three-dimensional space defined by horizontal and vertical position of the TD sensor and the horizontal position of the UL sensor. The distributions of the articulatory Euclidean distance values are summarised in Figure \ref{fig:art_ED}, depending on the item. The vowels in \emph{boy} and \emph{buy} have the greatest articulatory Euclidean distance, followed by \emph{bay}, \emph{bough}, \emph{beer}, \emph{beau} and \emph{boo}. In comparison, the  articulatory Euclidean distance is low for \emph{bee}, \emph{bore}, \emph{bar}, \emph{bear} and \emph{burr}. These observations are consistent with canonical diphthongs being characterised by greater articulatory displacement, compared to canonical monophthongs. We further note that the distribution of the articulatory Euclidean distance values is fairly continuous, except for a break separating two vowels, \emph{buy} and \emph{boy} from the rest. In addition, some vowels showed considerable variance in the distance values, notably \emph{beer}.

\subsection{Articulatory velocity} \label{art_vel}

\begin{figure}[htbp]
\begin{center}
\includegraphics[scale=0.8]{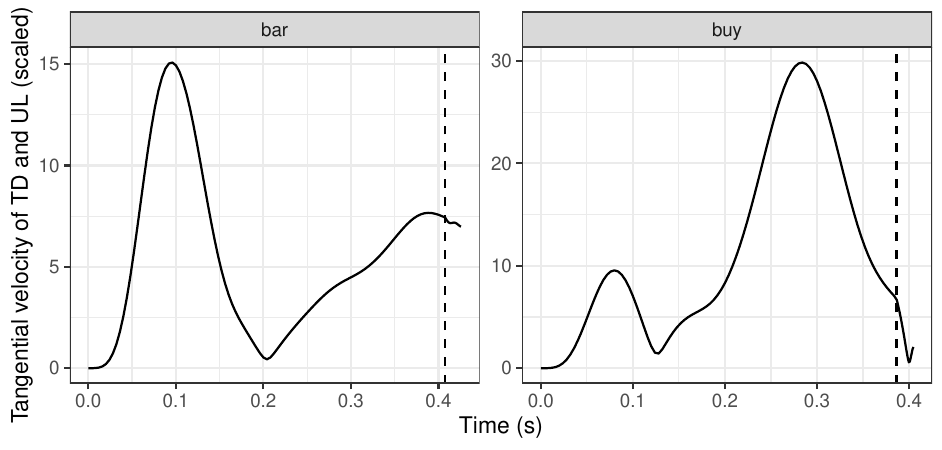}
\caption{TD-UL velocity for two example tokens pronounced by speaker f04. The dashed line represents the acoustic offset of the vowel.}
\label{fig:f04_real_ex}
\end{center}
\end{figure}

Let us now examine vowel velocity trajectories from the point of view of diphthongisation. Figure \ref{fig:f04_real_ex} illustrates the tangential TD-UL velocity profile for two representative tokens,  \emph{bar} and \emph{buy}), produced by speaker f04. In \emph{bar}, the velocity trajectory is characterised by an initial velocity peak, followed by a slow decline in velocity, and then another velocity rise. This type of trajectory is consistent with movement towards a single articulatory vowel target, followed by gestural release.

In comparison, \emph{buy}, is markedly different. For this velocity trajectory, we find an initial peak followed by a local minimum, but then velocity rises rapidly in the second half of the vowel, and another local minimum can be seen after acoustic offset of the vowel. This velocity profile is consistent with the vowel having two distinct articulatory targets, and most of the velocity profile is dominated by the movement towards the second target. 

\begin{figure}[htbp]
\begin{center}
\includegraphics[scale=1]{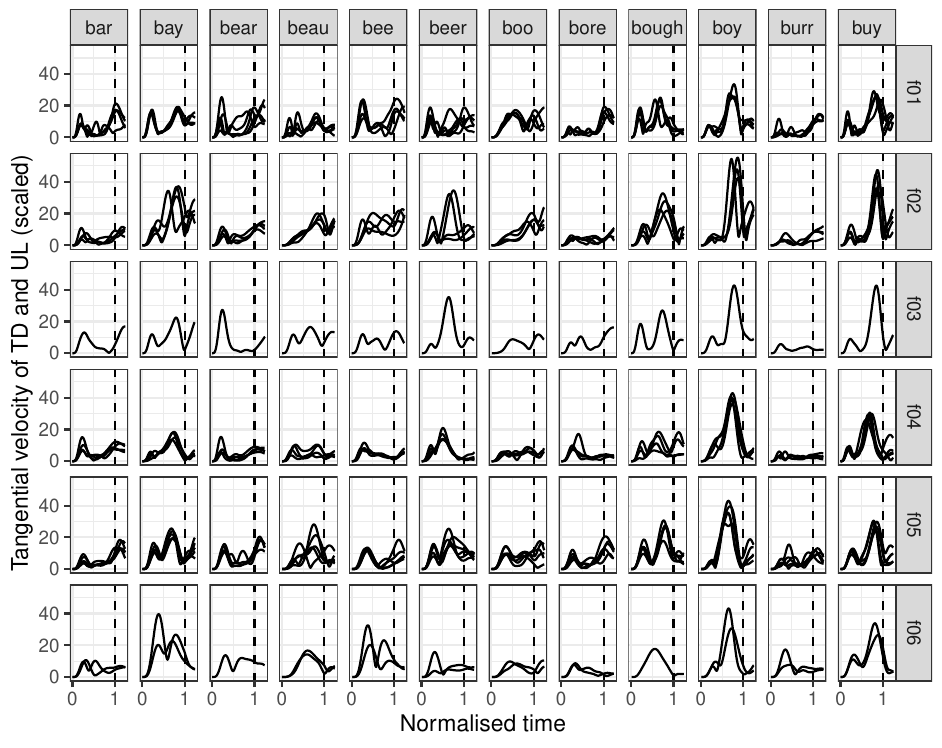}
\caption{TD-UL velocity profiles for all the individual vowel tokens}
\label{fig:art_vel}
\end{center}
\end{figure}

These two examples would suggest that at least some vowels have velocity profiles consistent with the hypothesis that monophthongs, like \emph{bar}, have a single gestural target, whereas diphthongs, like \emph{buy}, have two targets. Targets can be clearly discerned, corresponding to local velocity minima. However, the question is whether we can classify the velocity profiles of all vowels as having one or two targets. The velocities for the individual vowels are shown in Figure \ref{fig:art_vel}, plotted in normalised time and overlaid by speaker and by item. As we can see in the figure, many trajectories clearly align with one of the two types, however, the classification is not entirely straightforward in some cases. It is clear that the height of the component peaks can vary considerably, with some peaks being of limited height. However, there is also noise in the data, such that we see a series of small peaks in some cases. If the number of peaks were to be classified manually by a researcher, this would require some potentially arbitrary decisions about what does and what does not count as a peak. In order to avoid this problem, we instead undertook a systematic data-driven approach to parameterising the information about velocity trajectories, using a functional Principal Component Analysis (fPCA; \citealt{gubian2015}).

\begin{figure}[htbp]
\begin{center}
\includegraphics[scale=1]{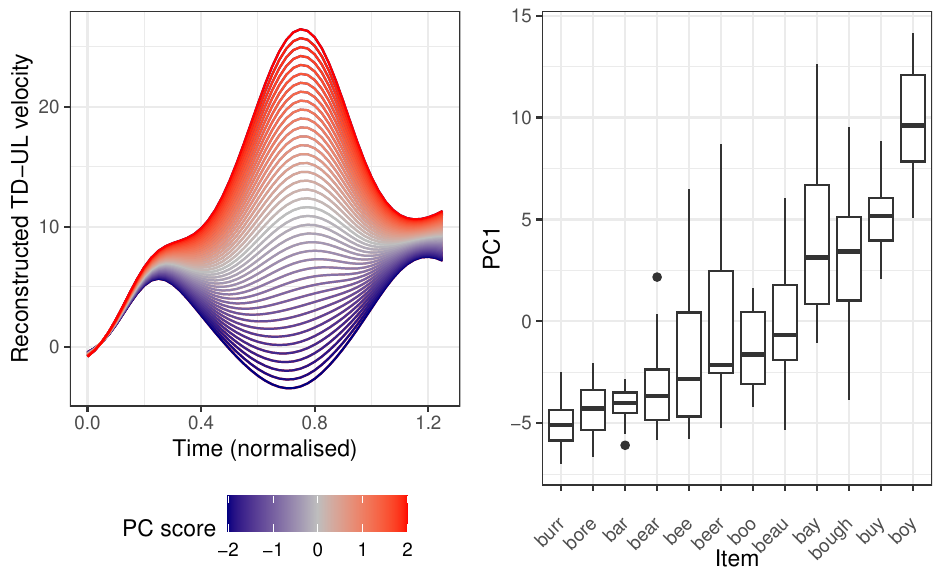}
\caption{Left: Perturbation of the articulatory velocity trajectory depending on PC1 score. Right: PC1 score depending on the item.}
\label{fig:pca_art}
\end{center}
\end{figure}

FPCA is a statistical method for reducing variance in time-varying measurements to orthogonal principal components. The tangential velocity profiles were input to the analysis for a time window corresponding to the acoustic duration of the vowel followed by a fixed 75ms window. Based on this analysis, 98\% of variance related to the spatio-temporal information in the velocity data can be reduced to four Principal Components. These components are illustrated in Appendix A. For the purpose of our analysis, we focus on the first Principal Component, PC1, which captured 61\% of the variance. The left panel of Figure \ref{fig:pca_art} shows a perturbation plot, which illustrates how variation in the PC1 score affects the shape of the velocity trajectory. The right panel of this figure shows the effect of item on PC1 score.

As we can see from the perturbation plot in Figure \ref{fig:pca_art}, most of the dynamic variance associated with variation in PC1 occurs around 0.75 of normalised vowel duration. An increase in PC1 corresponds to an increase of the velocity value at this time point, which is also the local velocity peak. Conversely, a decrease in PC1 score is associated with a lower velocity values at the same time point, creating a local velocity trough as PC1 falls further below 0. As a result of this perturbation,  two local velocity minima emerge in the corresponding velocity trajectory when PC1 is positive. From the right panel of Figure \ref{fig:pca_art}, we can see that such positive scores are typically found for \emph{bay}, \emph{bough}, \emph{buy} and \emph{boy}.
 In contrast, low PC1 scores correspond to velocity trajectories in which a single local minimum can be discerned. Cross-referencing with the right panel of Figure \ref{fig:pca_art}, such low scores are found for  canonical long monophthongs like \emph{burr}, \emph{bore} and \emph{bar}. Based on this interpretation, we can take PC1 as a proxy for diphthongisation. As we can see from the right panel of Figure \ref{fig:pca_art}, the distribution of the PC1 scores largely corresponds to an expectation that we might have about monophthongs and diphthongs. We find the highest PC1 scores for \emph{boy, buy, bough} and \emph{bay}, followed by \emph{beau, boo, beer} and \emph{bee}, and then the median PC1 scores tail off, and are low for \emph{bar, bear, bore} and \emph{burr}. Crucially, the distribution of PC1 is entirely continuous -- while it may not be unimodal, PC1 scores in all ranges are attested, suggesting an intermediate degree of diphthongisation for the vowels in the middle of the distribution, such as vowels in \emph{beau}, \emph{boo} and \emph{beer}.

\begin{figure}[htbp]
\begin{center}
\includegraphics{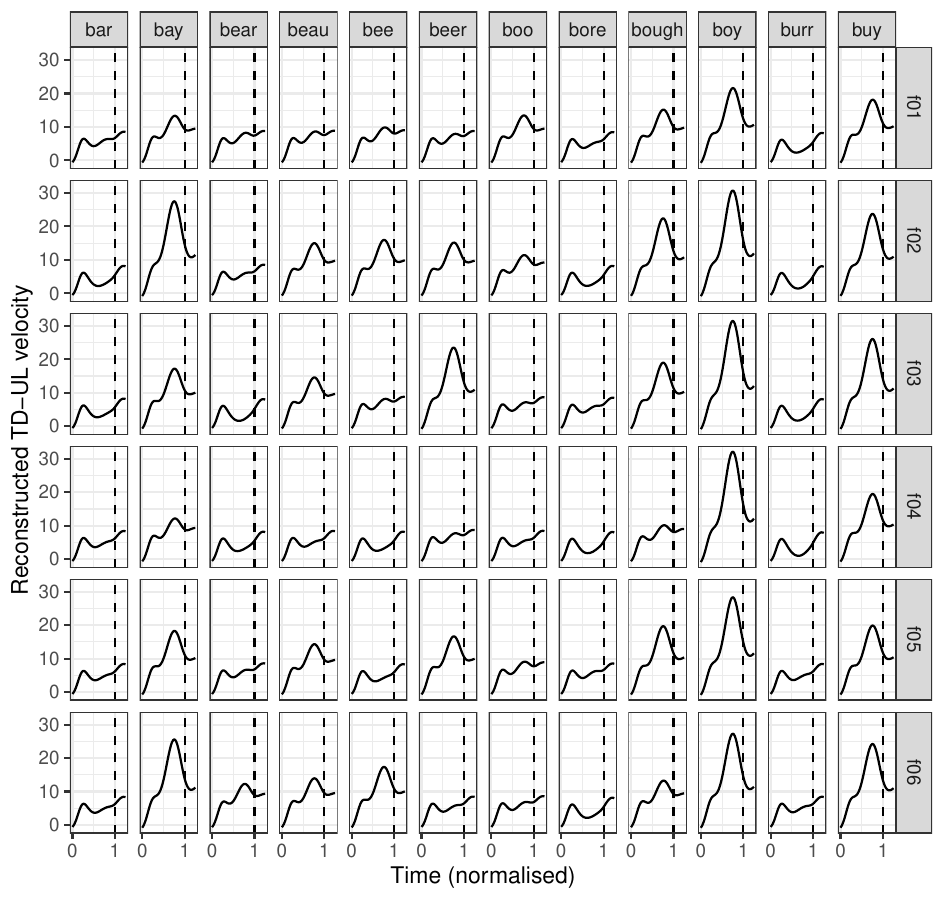}
\caption{Reconstructed TD-UL velocity trajectories based on by-vowel, by-speaker mean value of PC1}
\label{fig:pca_art_ind}
\end{center}
\end{figure}

Let us consider what the intermediate PC scores mean in terms of velocity trajectories. In principle, intermediate PC1 scores could arise from some vowels varying categorically between a one-minimum vs two-minima type of trajectory (categorical variation between a monophthong and a diphthong), or they could genuinely represent a trajectory that is intermediate between the two prototypical categories. In order to explore this question further, we reconstructed the velocity trajectories based on mean by-item and by-speaker PC1 scores, following the procedure in \cite{cronenberg2020}. The reconstructed trajectories, shown in Figure \ref{fig:pca_art_ind}, suggest a mixture of categorical and gradient variation in the trajectory shape, as captured by PC1. The vowels with intermediate PC1 scores, \emph{beau}, \emph{boo} and \emph{beer}, alternate between a monophthongal one-minimum trajectory with a trough in the second half and a diphthongal trajectory with two minima and a peak in the second half. For example, \emph{bee} is monophthongal for f04 and f05, but diphthongal for the remaining speakers. However, we also observe more gradient variation in the height of the second peak. This is evident in vowels with intermediate values of PC1, such as \emph{beer}, but also in canonical diphthongs like \emph{boy} or \emph{buy}. In this case, greater height of the second peak can be interpreted as a correlated of increased diphthongisation, related to greater distance between the targets within a diphthong.

\subsection{Formant displacement} \label{form_dis}
\begin{figure}[htbp]
\begin{center}
\includegraphics[scale=1]{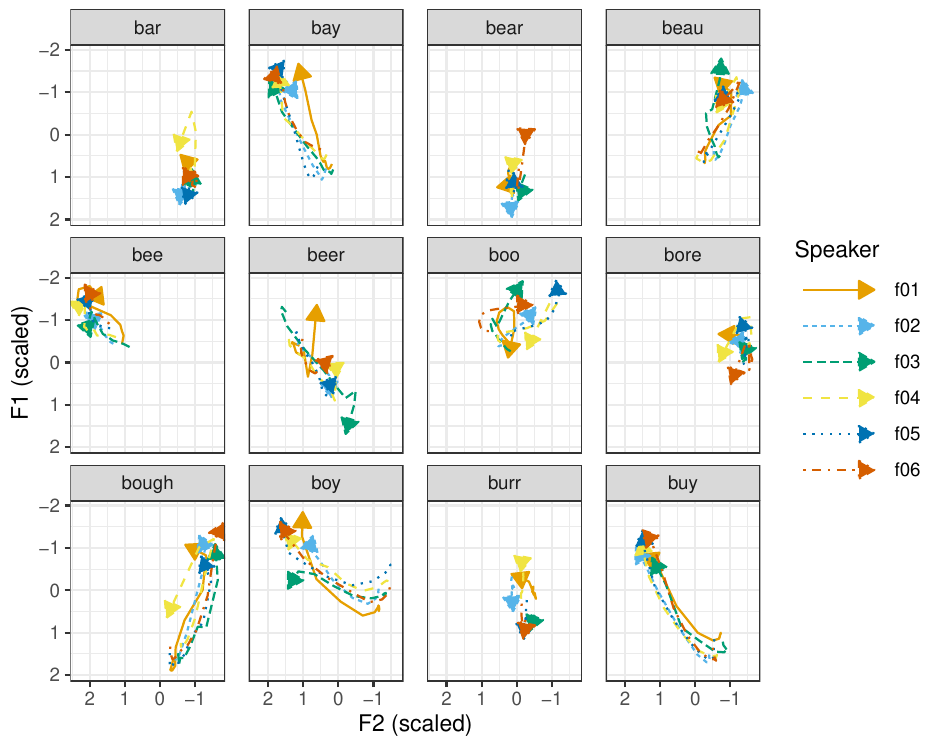}
\caption{Mean F1 and F2 displacement, depending on speaker and item.  The trajectories represent the mid 90\% of the vowel. }
\label{fig:form_disp}
\end{center}
\end{figure}

In order to compare our articulatory findings to a more familiar acoustic measure, we analysed the vowel formant trajectories.  Mean F1 and F2 trajectories for each speaker (GAM-smoothed) are plotted in Figure \ref{fig:form_disp}, depending on the item. The trajectories represent the mid 90\% of the vowel rather than the entire duration, due to difficulty of obtaining reliable formant measurements at the edges of the vowel.

\begin{figure}[htbp]
\begin{center}
\includegraphics[scale=1]{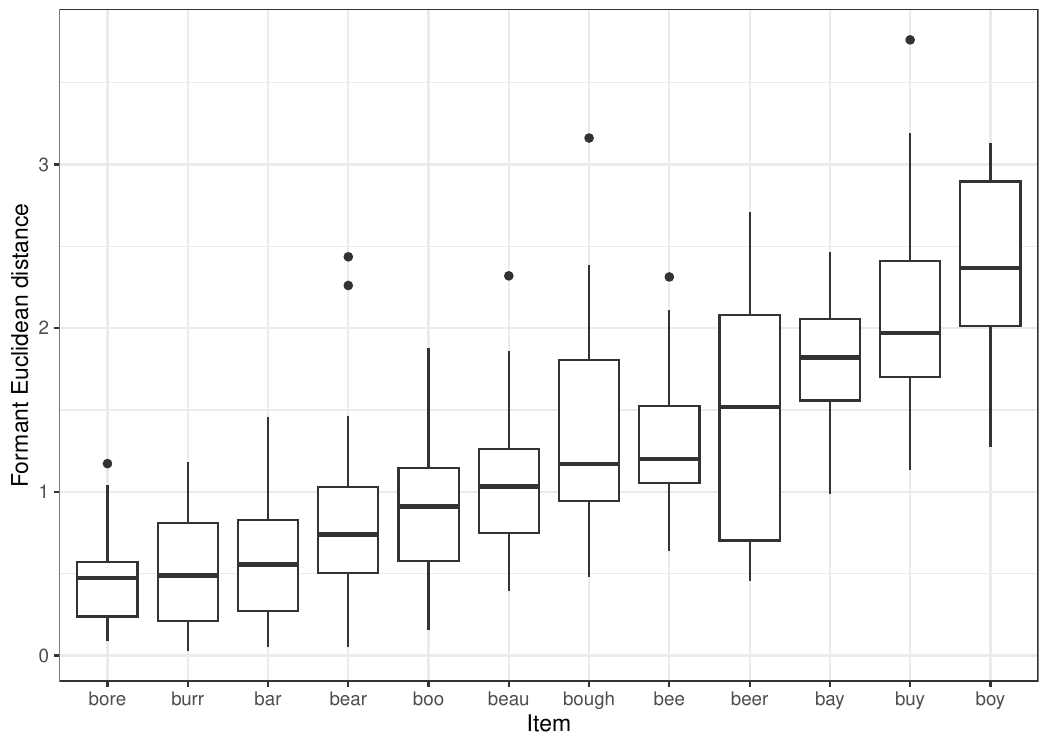}
\caption{Acoustic Euclidean distance (F1 and F2 displacement between 10 and 90\% of the vowel) by item}
\label{fig:form_ed}
\end{center}
\end{figure}

In general, we find a clear and pronounced formant excursion for \emph{buy, boy} and \emph{bough}, slightly lesser but still clear change for \emph{beau} and \emph{bay}. For \emph{beer}, \emph{boo} and \emph{bee}, there is variable intermediate degree of formant change, whereas \emph{bar, burr, bore} and \emph{bear} show very little change. These generalisations are broadly consistent with the Euclidean distance between formant values at 10 and 90\% of the vowel, as shown in Figure \ref{fig:form_ed}.

\subsection{Formant velocity} \label{form_vel}
In this part of the analysis, we focus on the rate of change in the formant trajectories, operationalised as tangential velocity of F1 and F2 displacement. The formant velocities for an example token of \emph{bar} and \emph{buy} are plotted in Figure \ref{fig:form_vel}. All the individual velocities are shown in Appendix B. In this case, we visualise the mid 90\% of the vowel trajectory in normalised time. 

\begin{figure}[htbp]
\begin{center}
\includegraphics[scale=1]{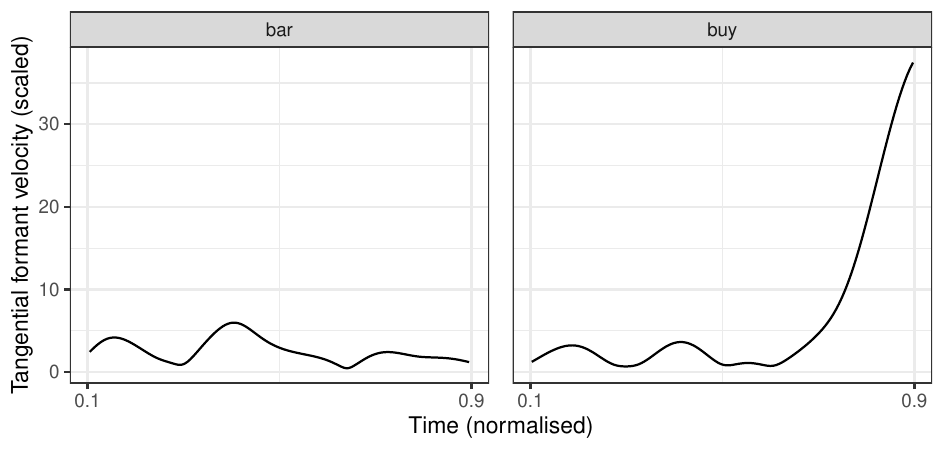}
\caption{Formant velocity profiles fortwo example tokens}
\label{fig:form_vel}
\end{center}
\end{figure}

The velocity profiles bear some similarities to the articulatory velocities presented in Section \ref{art_vel}. Notably, we can see a rapid rise in velocity in the second half of \emph{buy}. In contrast to articulatory data, the first half of the trajectory in this vowel does not show a clear initial peak. For \emph{bar}, the entire trajectory is relatively flat, with no discernible initial peak. The absence of such a peak is consistent with there being a steady state in the formants. 

\begin{figure}[htbp]
\begin{center}
\includegraphics[scale=1]{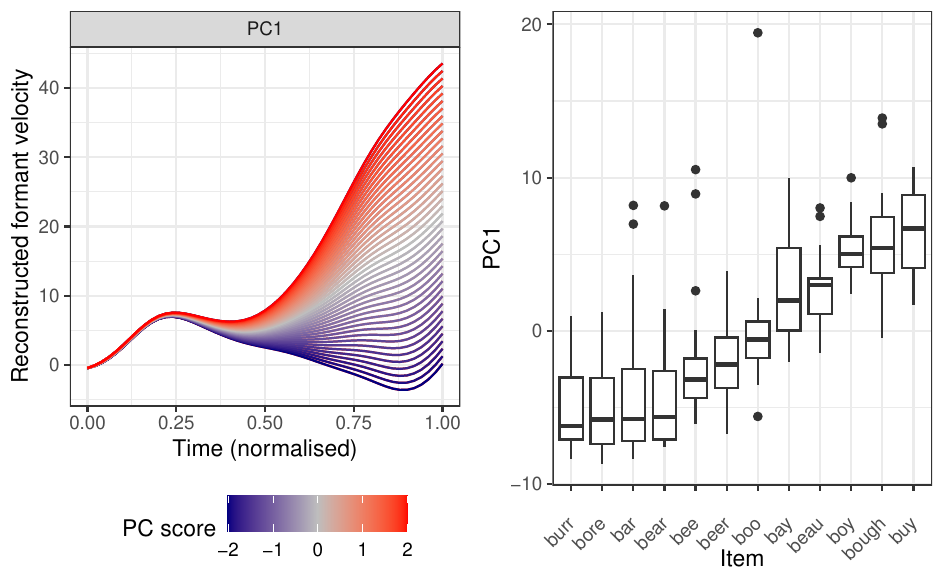}
\caption{Left: Perturbation of the formant velocity trajectory depending on PC1. Right: PC1 score depending on the item.}
\label{fig:pca_form_vel}
\end{center}
\end{figure}

We conducted an fPCA on the formant velocity data, in order to derive a numerical measure of formant change. The first Principal Component, illustrated in Figure \ref{fig:pca_form_vel}, captured 63\% of variance. As we can see in the perturbation plot in Figure \ref{fig:pca_form_vel}, PC1 is correlated with the presence of a peak in the second half of the vowel formant velocity trajectory. The higher the PC1 scores, the steeper the rise. Negative PC1 scores correspond to a trajectory that drops off slowly following an initial peak. We can generalise that PC1 is a measure of acoustic diphthongisation (presence vs. absence of a second velocity peak). Appendix C presents a perturbation plot for the first four PCs.

The right panel of Figure \ref{fig:pca_form_vel} shows the distribution of PC1 scores, depending on item. The relative values of acoustic PC1 scores closely resemble the results for articulatory PC1 (compare to Figure \ref{fig:pca_art}), whereas some small differences emerge between this measure and the formant Euclidean distance, shown in Figure \ref{fig:form_ed}. Specifically, \emph{beau} and \emph{bough} show more intermediate Euclidean distance values, whereas they are more  diphthongal, according to the acoustic PC1. The overall distribution of the acoustic PC1 is continuous, and intermediate values are represented.

\subsection{Hierarchical clustering} \label{clustering}
So far, it would seem that different measures of diphthongisation generally converge in distinguishing between vowels like \emph{boy} or \emph{buy} on the one hand, and vowels like \emph{bar} or \emph{burr} on the other. However, all the measures we have considered also yield intermediate values, with vowels in \emph{bee, boo, beau} and \emph{beer} tending towards the middle.

We performed a clustering analysis in order to explore whether any clusters of vowel emerge from the measures of diphthongisation that we have considered, and whether these clusters differ between articulatory and acoustic measures. We used hierarchical cluster analysing with Ward's clustering criterion \citep{murtagh2014, ward1963}. The input to clustering was a distance matrix based on by-vowel means for the four diphthongisation measures we have described: articulatory Euclidean distance, articulatory PC1, formant Euclidean distance and acoustic PC1. The correlations between these measures are included in Appendix D.

Figure \ref{fig:clusters} summarises the results of clustering, depending on different combinations of diphthongisation measures. Articulatory clustering was based on articulatory Euclidean distance and articulatory PC1. Acoustic clustering was based on formant Euclidean distance and acoustic PC1. Finally, a clustering was also performed on all the measures combined.

  \begin{figure}[htbp]
\begin{center}
\includegraphics[scale=1]{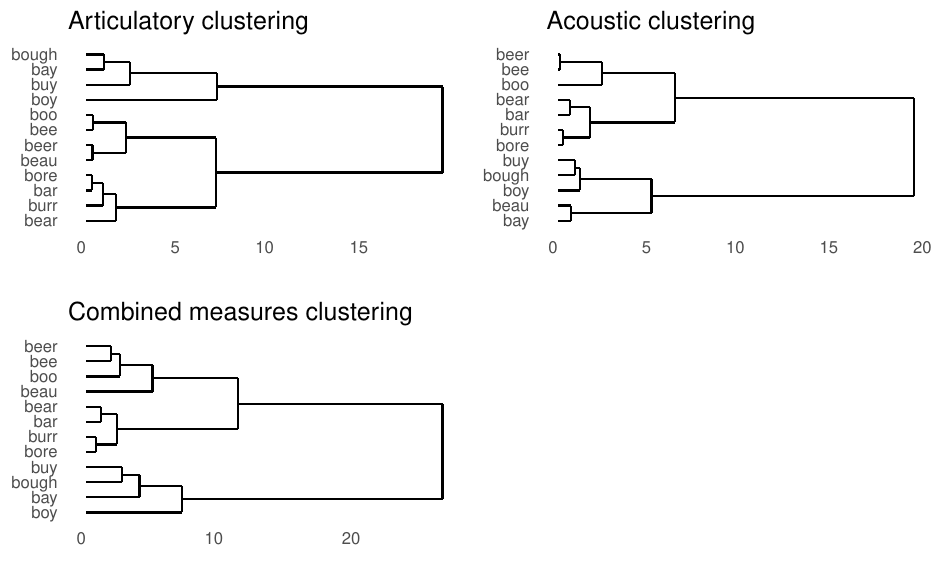}
\caption{Results of hierarchical clustering based on different measures of diphthongisation}
\label{fig:clusters}
\end{center}
\end{figure}

In all cases,  two to three clusters emerge from the data. The vowels in \emph{buy, boy, bough} and \emph{bay} always form a cluster. Additionally, the vowel in \emph{beau} is grouped with these vowels in the acoustics, but not in articulation, where it patterns with \emph{bee}, \emph{boo} and \emph{beer}. It is difficult to determine whether this vowel presents a case where articulatory displacement and acoustic change diverge systematically, or whether the discrepancy is an artefact of the measures we used. Another cluster is formed consistently by vowels \emph{bar}, \emph{burr}, \emph{bore} and \emph{bear}. A third cluster is formed by \emph{bee}, \emph{beer}, \emph{boo} and (only in articulation) \emph{beau}. 

\begin{figure}[htbp]
\begin{center}
\includegraphics[scale=1]{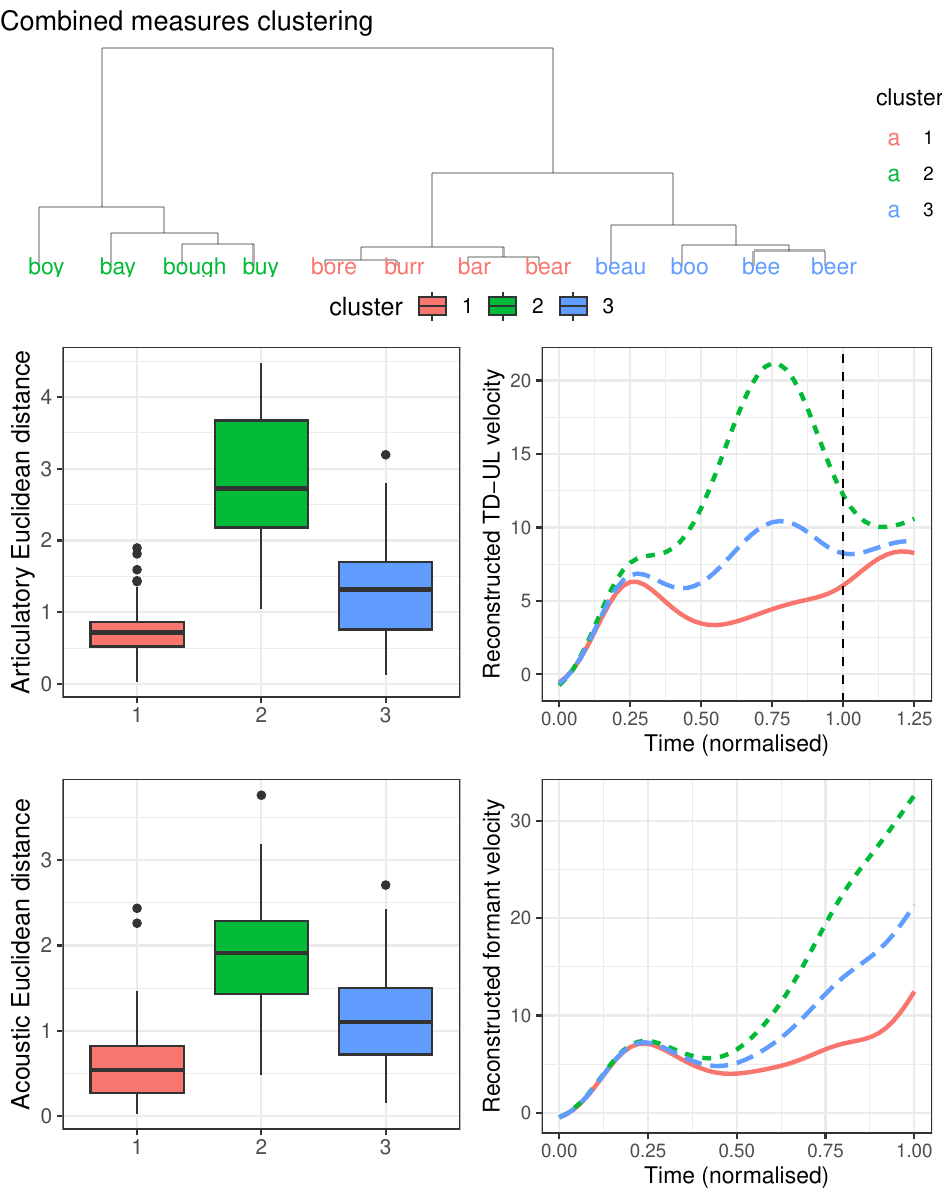}
\caption{The effect of cluster on the four measures of diphthongisation}
\label{fig:preds}
\end{center}
\end{figure}

Based on previous analysis, we may have an expectation of the dynamic properties of the three clusters, as showing different degrees of diphthongisation. In order to verify them, we plotted the four measures of diphthongisation as a function of cluster. The relevant plots are in Figure \ref{fig:preds}. Combined measures clustering was taken as the basis for these plots. For the articulatory and acoustic Euclidean distances, we plotted their values by cluster. For the articulatory velocity and the formant velocity, we calculated the by-cluster mean values of the articulatory PC1 and acoustic PC1, and we reconstructed the relevant velocity trajectories for each cluster, based on these means.

Cluster 1 comprises the vowels \emph{bar, burr, bore, bear}. These are are canonical monophthongs for which we can see a single articulatory target, a formant velocity consistent with a steady state (little formant change in the second part of of the vowel) and a small degree of articulatory and acoustic displacement. Cluster 2 is formed by \emph{buy, boy, bough} and \emph{bay}. We can interpret this group as canonical diphthongs that show a two-target articulatory velocity, a clear late peak in formant velocity and a large degree of articulatory and acoustic displacement. Cluster 3 includes vowels \emph{bee, boo, beau} and \emph{beer}. They form an in-between category characterised by some articulatory and acoustic displacement, as reflected by the values of the Euclidean distances. The average TD-UL velocity trajectory for this cluster shows the presence of a second velocity peak, which is however limited in height. However, to some extent, the intermediate nature of this category is due to inter-speaker  variation. Comparing Figure \ref{fig:pca_art_ind}, vowels like \emph{bee}  and \emph{beer} may show a single peak for some speakers, but two peaks for other speakers. Acoustically, this produces an intermediate mean degree of change: greater than in canonical monophthongs, but less than in canonical diphthongs.

\subsection{Summary of the results}
In response to our research questions, two key observations emerge from our combined articulatory and acoustic analysis of diphthongisation in vowels. Firstly, for some vowels, (e.g. \emph{buy}, \emph{boy}), we can very clearly discern two distinct articulatory targets, consistent with the predictions of a compositional model of diphthongs. At the other end of the scale, some vowels (e.g. \emph{burr}, \emph{bar}) only show one discernible articulatory target. Secondly however, we observe that it is not possible to draw a boundary between two-target and one-target vowels. All the potential measures of diphthongisation that we have considered show gradience, and they commonly include intermediate values that fall between canonical diphthongs and canonical monophthongs. Such intermediate values are typical of vowels in \emph{bee}, \emph{boo}, \emph{beer} and \emph{beau}. These vowels can vary between a one-target and two-target trajectory type, depending on the speaker, but they also show more gradient variation, with varying height of the second velocity peak. This type of trajectory corresponds to intermediate degrees of articulatory displacement and acoustic formant displacement. 

To account for these facts, a representational model must reconcile some aspects of categoricity and gradience. The key question for articulatory modelling is how to capture gradient articulatory variation between one-target and two-target vowels. While the phonetic manifestation of the variation shows gradience, the underlying number of targets is not continuous: an articulatory target is either present or absent. We must therefore consider which articulatory parameters can give rise to the kind of variation we find. We address this question through simulation, using a task dynamic model of gestural coordination \citep{saltzman1989, browman1992, sorensen-gafos2016}.

\section{Computational modelling} \label{model}

\subsection{Aims}

This section presents simulations of articulatory dynamics in vowels using a task dynamic model. This allows us to generate quantitative predictions based on an explicit set of theoretical assumptions. In doing so, we can test the predictions of the AP/TD model by comparing the output of the simulation to the empirical data (see \citealt{burgdorf-tilsen2021, hsieh2017, marin2007} for a similar approach for developing AP models of vowels). Our specific aim is to evaluate the following two proposals that arise from earlier theoretical proposals discussed in Sections \ref{comp}--\ref{theoretical_issues}. 

\begin{enumerate}[i.]
\item Diphthongs have two targets, and long monophthongs have a single targets (i.e. diphthongs are compositional, and monophthongs are not). In evaluating this proposal we focus especially on the issue of vowel duration, and discuss several possible mechanisms to capture the relevant dynamic and duration facts.
\item A two-target model of all long vowels (i.e. both diphthongs and monophthongs are compositional). In this view, a long monophthong comprises two (near-)identical targets and a diphthong comprises two distinct targets. The difference between a long monophthong and a diphthong can, therefore, be modelled as a gradient change in a target's gestural parameters. This type of model has been sketched out in previous literature, but it has not been shown that a two-target model can generate realistic articulatory data, especially for monophthongs and for gradient diphthongisation.
\end{enumerate}

\subsection{Method}

We simulated gestural dynamics using Equation \ref{sg16}, which is a modified task dynamic model proposed by \citet{sorensen-gafos2016}. This differs from the classic \citet{saltzman1989} model in the addition of a cubic term $dx^{3}$, which acts as a non-linear restoring force on the spring in the damped mass-spring model. This corrects for the overly short time-to-peak velocity and asymmetric  velocity profiles in previous models. Values of $d > 0$ increase the strength of the non-linear restoring force. In our simulations, a uniform value of $d$ adequately reproduces the qualitative velocity distinctions that are central to our predictions, but we here optimise $d$ separately for each simulation in order to show that the model is also capable of generating empirically-realistic velocity profiles. Additionally, $b$ is a damping parameter defined as $b = 2\sqrt{mk}$, where $k$ is stiffness and $m$ is a mass parameter that is always equal to 1.

\begin{equation}
\label{sg16}
\ddot{x} + b\dot{x} + kx - dx^{3} = 0
\end{equation}

In all models, we simulate the tract variable Tongue Body Constriction Degree (TBCD) as a proxy for the tongue dorsum constriction, which we define in the normalised coordinate space [0,1], where $x =$ 0 is minimally constricted (representing a maximally open constriction) and $x =$ 1 is maximally constricted (representing complete closure) \citep{burgdorf-tilsen2021}. In all cases, we assume all vowel gestures to have uniform stiffness ($k$). Note that this is not an essential feature of our model, as variation in stiffness can form part of gestural representation, but we do this to show that our predictions do not intrinsically rely on stiffness variation.

The timing between two vowel gestures was defined using a coupled oscillator model of gestural coordination in Equation \ref{osc} \citep{tilsen2018}. $\Phi_{ij}$ is the relative phase between oscillators $i,j$, such that $\Phi_{ij} = \theta_{i} - \theta_{j}$. $C_{ij}$ is a matrix of coupling strengths between oscillators $i,j$, where $C_{ij} > 0$ is in-phase and $C_{ij} < 0$ is anti-phase.

\begin{equation}
\label{osc}
\dot{\theta}_{i} = 2\pi f_{i} + \Sigma_{j} C_{ij} \sin(\Phi_{ij})
\end{equation}

We make the simplifying assumption that a vowel gesture is defined as a 250 ms period of gestural activation and all gestures have the same oscillator frequency $f$ = 4 Hz. Anti-phase coupling between two vowel gestures results in a 125 ms lag when oscillator frequencies are 4 Hz, so a vowel with two 250 ms gestures coupled anti-phase is 375 ms in duration. The 250 ms duration of gestural activation intervals were set by hand based on the average pattern in the empirical data. This is a simplifying heuristic for the purposes of illustration; a feedback-based suppression mechanism is instead a more likely approximation of how speakers control gestural deactivation and, therefore, the timing of gestural offsets \citep{tilsen2022}.

All simulations were conducted using the Python programming language. We solved for the velocity of TBCD using the Explicit Runge-Kutta method via SciPy’s \texttt{integrate.solve\_ivp} function \citep{SciPy-NMeth2020}. In all cases, we solve using a time step of $\Delta t$ = 0.001 seconds and all simulations have uniform stiffness ($k$ = 2000) and damping ($b = 2\sqrt km$, where $m$ = 1). The value of $d$ is manually specified for each gesture to produce velocity trajectories similar to those observed in empirical data. Further details can be found in the online documentation at \url{https://osf.io/gub32/}.

\subsection{Modelling one-target monophthongs vs. two-target diphthongs}

We begin by simulating canonical examples of a monophthong and a diphthong, represented by \emph{bar} and \emph{buy} respectively. We selected these examples, because \emph{bar} consistently shows a single velocity minimum for all speakers in our data, whereas \emph{buy} consistently shows two minima. In addition, the vowels in \emph{bar} and \emph{buy} are unrounded, which allows us to focus on TBCD as a primary correlate of articulatory movement inherent to the vowel. We simulated the TBCD velocity trajectories for \emph{bar} and \emph{buy} assuming the former consists of a single dorsal gesture (TBCD = 0.3), and the latter consists of two distinct dorsal gestures coupled anti-phase (TBCD = \{0.3, 0.9\}), as hypothesised in Section \ref{intro}.

\begin{figure}[htbp]
\begin{center}
\includegraphics[scale=0.6]{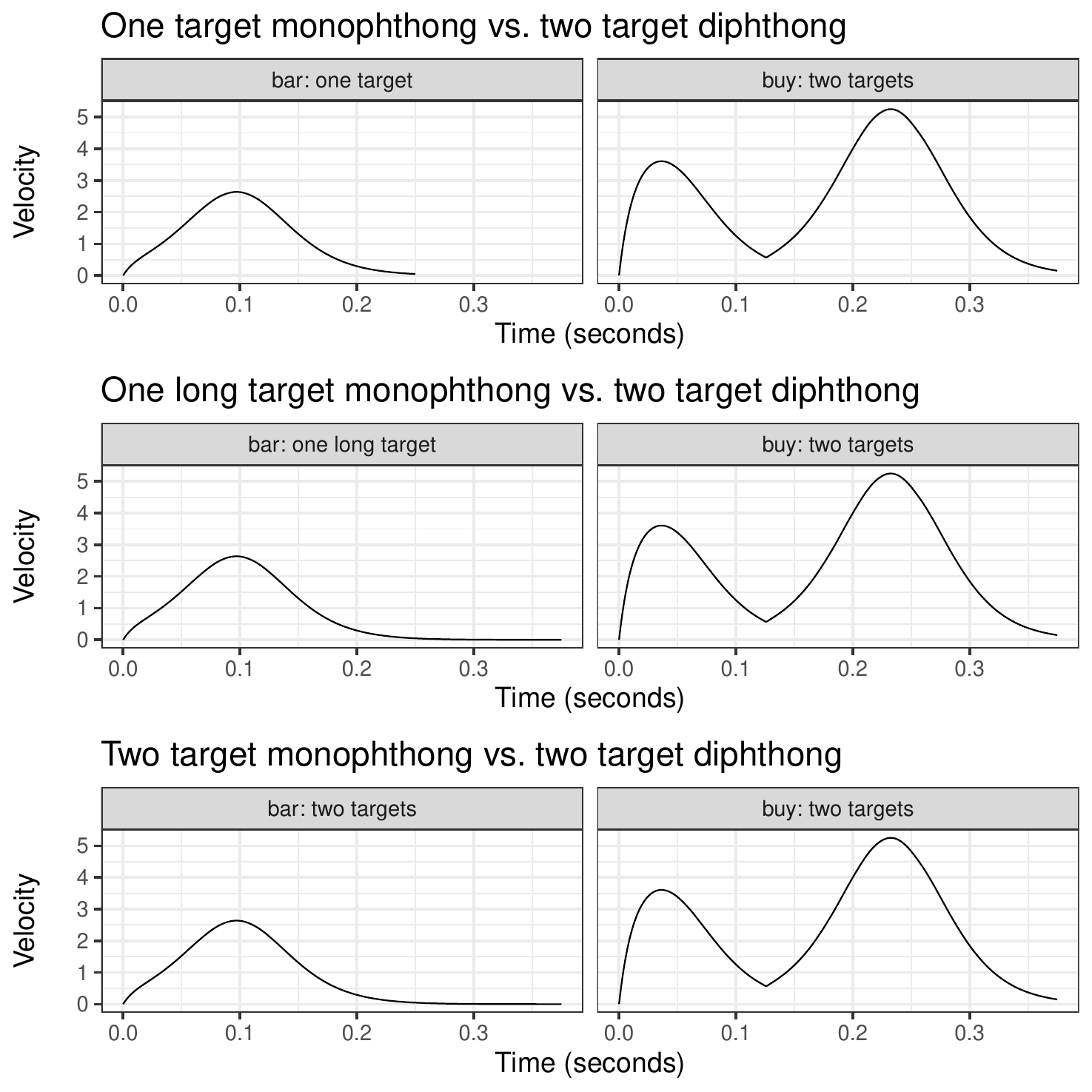}
\caption{Simulated TBCD velocities for \emph{bar} and \emph{buy} under three sets of assumptions}
\label{fig:sim_combined}
\end{center}
\end{figure}

The top panel of Figure \ref{fig:sim_combined} shows the results of the simulation. The TBCD velocity trajectories pattern as expected in terms of shape: the one-gesture trajectory shows a single peak followed by a local minimum, whereas the two-target trajectory consists of two velocity peaks. Note that there is no final gestural release in the simulated data. Notably, the model predicts a durational difference between the two vowels: the one-target monophthong is shorter than the two-target diphthong (250 ms vs. 375 ms respectively).

\begin{figure}[htbp]
\begin{center}
\includegraphics[scale=1]{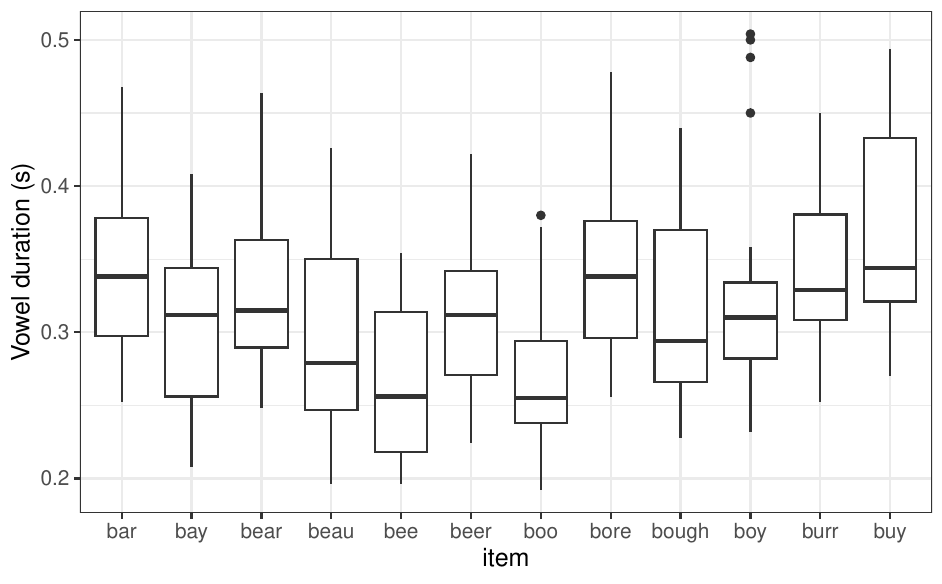}
\caption{Vowel duration by item}
\label{fig:durs}
\end{center}
\end{figure}

In general, diphthongs have not been reported to be longer than monophthongs in English \citep{peterson1960, lehiste1961}. Nevertheless, duration ought to be investigated more systematically in light of the modelling outcome. Figure \ref{fig:durs} shows the distributions of vowel duration, depending on the item. As we can see, the duration of the vowel in \emph{buy} is not systematically longer than that in \emph{bar}. More generally, it is also not the case the canonical diphthongs are longer than canonical monophthongs. While there is some variation in vowel duration, it mainly seems correlated with vowel height, in line with higher vowels being inherently shorter than low vowels \citep{peterson1960}.

Let us consider some alternative models that do not predict a duration difference between long monophthongs and diphthongs. One possibility is to model the two component gestures in diphthongs as being coupled in-phase. As previously discussed in Section \ref{comp}, in-phase coupling has previously been proposed for ongliding diphthongs, such as /\textipa{ju:}/, in contrast to offgliding diphthongs such as \textsc{price} \citep{marin2007, hsieh2017}. A key difference between these two sets of vowels is syllable weight: the glide does not contribute to syllable weight in ongliding diphthongs, but it does so in the case of offgliding diphthongs. This can be captured through a coupling asymmetry between the two sets of vowels that mirrors the weight asymmetry between onsets and codas: onset consonants, coupled in-phase to the vowel, are transparent to syllable weight, whereas coda consonants, coupled anti-phase to the vowel, carry syllable weight. Thus, modelling diphthong gestures as coupled in-phase predicts that diphthongs are phonologically light, which is incorrect for offgliding diphthongs such as \emph{buy}.

An alternative proposal is that long monophthongs have a single long gesture, whereas diphthongs are a composition of two short gestures. In this view, both classes can have similar duration, but it arises from different sources. To illustrate this, the middle panel of Figure \ref{fig:sim_combined} shows \emph{bar} modelled as a single long target at TBCD = 0.3, with 375 ms duration. This model seems to provide a good empirical fit, generating a velocity trajectory that has the desirable dynamic properties (presence of a single velocity peak) and the expected duration equivalent to diphthong vowels.

\subsection{Modelling long monophthongs as two-target vowels}

Let us now consider whether realistic articulatory vowel dynamics can be generated by a model in which diphthongs are modelled as comprising two targets coupled anti-phase, and the same is true for long monophthongs. In case of long monophthongs, the component targets are identical in terms of gestural constriction and location, whereas for diphthongs, the constriction and location of the component targets differ, resulting in inherent vowel change. This is in line with a previous proposal by \cite{popescu2022}. The main modification we propose is that all phonologically long vowels in English have two articulatory targets, and not just the vowels characterised by some degree of audible or measurable diphthongisation. Note that in this model we do not consider glides to have a separate status and instead model diphthongs as the composition of two short monophthong targets.

Example simulations of \emph{bar} and \emph{buy} are in the bottom panel of Figure \ref{fig:sim_combined}. Specifically, \emph{bar} was modelled as having two vocalic gestures, with the same TBCD targets for each gesture (TBCD = \{0.3, 0.3\}), which were coupled anti-phase to one another (we refer to the first as the nucleus and the second as the offglide). Note that the gestural parameters of \emph{bar} are identical to one-target \emph{bar} in the middle panel of Figure \ref{fig:sim_combined}, except here we have two monophthongal gestures coupled anti-phase, rather than a single long gesture. The vowel in \emph{buy} has two vocalic gestures with different targets (TBCD = \{0.3, 0.9\}). The blending ratio was set at 1:100 in favour of the offglide. As we can see, a monophthongal trajectory with a single local minimum (a single target) emerges from two underlying identical targets, but the duration of the monophthong increases, matching the predicted duration of a diphthong.

So far, we have seen that the two-target model produces a plausible velocity trajectory for canonical monophthongs, and it correctly predicts that diphthongs and long monophthongs are phonologically heavy and have a similar phonetic duration. We now further consider whether such a model can also capture intermediate degrees of diphthognisation and variable diphthongisation, characteristic of vowels such as 
in \emph{bee}.

We modelled the long monophthong /i/ comprising two targets, which we refer to as the nucleus [i] and glide [j]. \citet{burgdorf-tilsen2021} show that the differences between high vowels and glides can result from syllable organisation (vowels occur as nuclei; glides occur as onsets/codas) rather than different gestural specifications, but we also allow for the possibility of variability between them. Specifically, we explore the possibility that variation in diphthongisation of /i/ arises through variation in the nucleus TBCD target, with no variation in other articulatory parameters. 

First, we fix the TBCD target for [j] at 0.9, representing a palatal constriction, but allow the TBCD target for [i] to vary across the range \{0.9, 0.8, 0.7, 0.6\}. For example, a TBCD target of [i] = 0.9 and [j] = 0.9 gives identical targets for nucleus and offglide, whereas a value of [i] = 0.6 and [j] = 0.9 represents a more open constriction for the nucleus than the glide. As in our previous simulations of two-target vowels, we use a blending ratio of 1:100 in favour of the offglide and uniform stiffness for both gestures. We note that the blending ratio and stiffness values are not a requirement of our model and qualitatively similar results can be obtained using lower blending ratios or different stiffness values for nucleus and coda.

\begin{figure}[htbp]
\begin{center}
\includegraphics[scale=0.6]{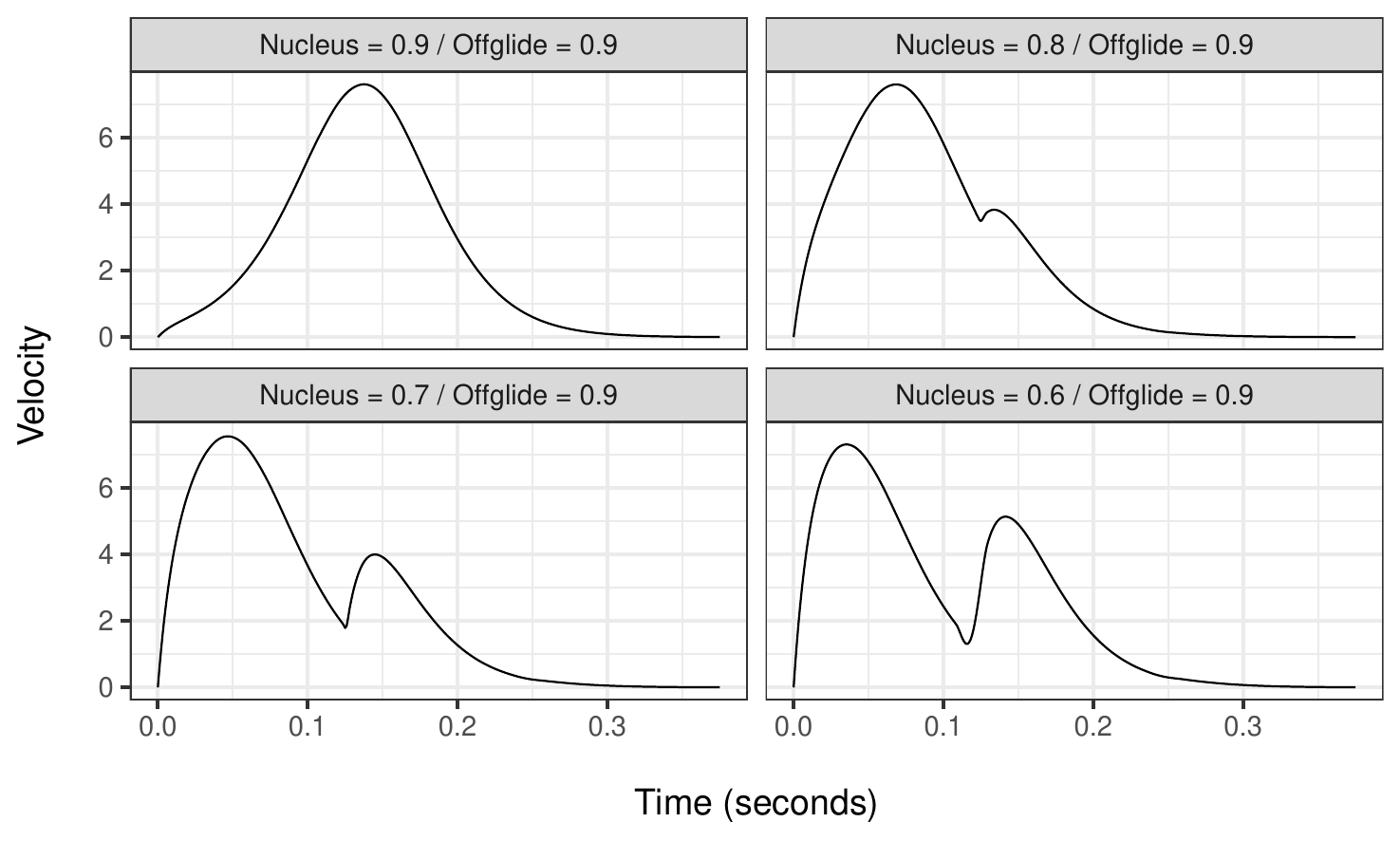}
\caption{Simulated TBCD velocity for \textit{bee} across different [i] TBCD target values. Nucleus = 0.9 corresponds with a classic one-target long monophthong. In all cases, the offglide target is TBCD = 0.9.}
\label{fig:fleece_targets}
\end{center}
\end{figure}

Figure \ref{fig:fleece_targets} shows that when TBCD targets are identical (TBCD = 0.9) the model produces a one-target velocity profile for /i/. Changing the nucleus TBCD target to a lower (more open) value produces a successively larger second velocity peak, demonstrating that /i/ becomes more diphthongal as the nucleus target diverges further from the offglide target. The variation in the height of the second velocity peak is qualitatively similar to the variation observed in the empirical data. As shown in Figure \ref{fig:art_vel}, individuals vary in their production of \emph{bee} with respect to how high the second peak is. For some of them, there is no second peak, whereas for others, the second peak is present, but it is not as high as in prototypical diphthongs. The model confirms that this variation can arise from varying the gestural constriction of the nucleus, while the remaining parameters remain unchanged. This would be impossible with the `one long target' model, without recourse to an additional mechanism for adding targets or splitting an existing target.

\subsection{Summary of modelling}

In sum, there are different ways of modelling the distinction between canonical diphthongs and canonical monophthongs. For diphthongs, a compositional model with two component targets generates two velocity peaks, same as we find in the empirical data. For monophthongs, a single velocity peak can be modelled either using a single long vowel target, or as a combination of two identical component targets. These two representations yield the same empirical predictions for canonical cases, and so modelling long monophthongs as one long target versus two coupled targets cannot be distinguished from the data alone. Nevertheless, we propose that the two-target modelled is preferred on theoretical grounds. We discuss these in Section \ref{discussion} below, with special attention paid to the case of gradient diphthongisation, which can be captured using two similar targets, as shown by our modelling.

\section{Discussion} \label{discussion}

In this study, we analysed the articulatory nature of diphthong production, in comparison to long monophthongs. In doing so, we have focused on two questions. The first question was whether diphthongs can be consistently analysed as comprising two articulatory targets, in contrast to monophthongs that could be characterised as comprising one target. Diphthong compositionality, the idea that diphthongs have two articulatory targets, finds support in the articulatory data. The movement of the articulators, as we have observed it, can be characterised as movement from one target to another. The key empirical reflection of the two targets in the abstract representation is the presence of two articulatory maxima (local velocity minima) in the dynamic trajectories of tongue dorsum movement and lip protrusion. Note that the second target may not always be reached, as proposed by \cite{lindblom1967}, and as confirmed here by tokens in which the second target is delayed beyond the acoustic offset of the vowel. 

The second question concerns the nature of the distinction between diphthongs and monophthongs. Traditionally, compositional diphthongs are represented in opposition to non-compositional monophthongs, which entails that the two are distinct phonological categories,  as discussed in Section \ref{comp}. Contrary to this prediction, we find that diphthongisation is gradient in the articulatory and acoustic domain. The clustering analysis in Section \ref{clustering} shows that regardless of whether we consider articulatory or acoustic measures of diphthongisation, an intermediate category arises between canonical monophthongs and canonical diphthongs, and this category includes the vowels with variable or limited degree of diphthongisation: \emph{bear, bee} and \emph{boo}. The observation is not trivial considering non-linearities between articulation and acoustics \citep{stevens1989}, which could produce gradient acoustic change from categorical articulatory shifts. While this is of course still possible in some cases, not all instances of intermediate diphthongisation can be explained in this way.

Modelling presented in Section \ref{model} shows that the articulatory properties of canonical diphthongs are well captured in a compositional model in which diphthongs consist of two articulatory targets coupled anti-phase. A two-target model is also well-suited to capturing the phenomenon of gradient diphtongisation. Our simulation shows that the difference between a long monophthong with no discernible diphthongisation and one with variable diphthongisation can be modelled as a change in the nucleus TBCD target (Figure \ref{fig:fleece_targets}). As a consequence, small changes in the degree of diphthongisation emerge in the system from variation in simple gestural parameters, with no change in the underlying structural organisation of gestures. When it comes to modelling long monophthongs, we seem to have two empirically equivalent alternatives. The dynamic and durational properties of long monophthongs can be accurately predicted by a model in which long monophthongs consist of two identical articulatory targets. However, they are equally well predicted by a single-target model assuming long target duration for monophthongs. Even though empirically, these models are equivalent, the two-target model is arguably preferred from the point of view of phonological theory, as well as from the patterns of variation and change.

A compositional two-target model of long monophthongs is theoretically appealing in establishing a systematic correspondence between phonological structure and syllable weight. Like diphthongs, but in contrast to short monophthongs, long monophthongs are phonologically heavy. The weight distinction between long and short monophthongs can be captured in our model using the tools independently proposed for coda consonants. In case of coda consonants, syllable weight follows from the presence of a consonant gesture coupled anti-phase to the vowel gesture. In a two-target model of long monophthongs, the structure is much the same, except the anti-phase coordination holds between two component vowel gestures, rather than a vocalic and a consonantal one. In contrast to long monophthongs, we can model short monophthongs as having only a single underlying target. An added benefit is that the difference in phonological structure and phonotactic behaviour of long and short monophthongs is in this case also systematically correlated with phonetic duration: one-target monophthongs are phonetically shorter than two-target monophthongs. 

Furthermore, while it is possible to model long monophthongs as a single long target, this requires additional assumptions, such as long monophthongs having an inherently different phonological specification from all other vowels. For example, one conceptualisation of activation interval duration is by mapping target achievement to a particular phase of a virtual oscillator cycle, such as 270$^\circ$ \citep{browman1995gestural}, with differences in duration arising as a consequence of oscillator frequencies (see \citealt{tilsen2018} for further illustration). A related conceptualisation of duration is time-to-target achievement, which is modulated by stiffness \citep{ratko-etal2023}. In both cases, long monophthongs must have a different gestural specification from short vowels and diphthongs. While not inherently problematic, this requires extra assumptions that must be justified over a simpler model.

 Timing mechanisms in the standard model of AP/TD have come under intense critique \citep{turk-shattuck-hufnagel2020}, so we also comment on the issue of long monophthongs versus diphthongs in light of newer developments in AP/TD. Selection-coordination theory poses a feedback-induced suppression mechanism for the timing of gestural offsets, whereby speakers learn to use external feedback to control gestural duration \citep{tilsen2016}. This could facilitate different gestural durations very easily, as speakers can use feedback in different ways to suppress a gesture depending on the communicative demands. However, this framework still supports a two-target model for long vowels: \citet{tilsen2016} explicitly states that long vowels are the result of `intentional reselection' of the same gesture and provides a developmental explanation for the emergence of such patterns. By extension, the gradient transition from a long monophthong to a diphthong can be viewed as gradual dissociation of the two component gestures, such that one gesture gradually takes on a different target value from the other gesture.

Having considered the phonological issues, let us now turn to the predictions made by the one-target model in terms of variation and change. There is no way for the one long target to become a two target vowel (i.e. undergo diphthongisation), without the categorical addition of another target, or some sort of `splitting' of the long target. If this is the case, then diphthongisation of long vowels should be somewhat constrained, because it requires a structural reorganisation, but the evidence presented in Section \ref{theoretical_issues} suggests that gradient diphthongisation is pervasive in English, both synchronically and diachronically. In comparison, the two-target model can produce gradient change from a long monophthong to a diphthong via variation in only the nucleus target value. This affords a historically accurate model of the gestural representation of English vowels. As shown for the \emph{bee} example in Figure \ref{fig:fleece_targets}, variation in the degree of diphthongisation can be captured through variation in the spatial properties of one of the component vowel targets -- it represents variation in vowel quality, rather than variation in underlying vowel structure. Variation in vowel quality is ubiquitous, and from that point of view, we might expect that variation in the degree of diphthongisation is also common. This expectation is consistent with observed variation in English. In the Northern English data presented here, there are several vowels that vary between more monophthongal and more diphthongal realisations. This is the case for \emph{bee, boo} and \emph{beer}, and the same mechanism would account for variation between monophthongal \emph{bay} and \emph{beau} for some Northern English speakers in contrast to a more diphthongal pronunciation in the pan-regional standard. Similar variation also exists in other varieties of English (e.g. Standard Southern British English; \citealt{lindsey2019}).

Following the same reasoning, historical changes such as diphthongisation and monophthongisation fall under a wider type of changes in vowel quality. In contrast, changes affecting vowel length are distinct, as they require major phonological reorganisation in terms of inserting or deleting a vowel target. The history of English illustrates a potential diachronic connection between phonological length and the presence of an underlying articulatory gesture. In non-rhotic varieties of English, many of the long monophthongs emerged historically from sequences of a short vowel followed by /r/. A plausible diachronic account of the changes is as follows. The coda /r/ initially involves a combination of a vocalic and a consonantal gesture, as is a common characteristic of liquids \citep{proctor2011, proctor2019}. The final liquid then undergoes a gestural reorganisation in which the consonantal gesture is gradually reduced. The vocalic gesture of the liquid, however, remains in place, becoming a diphthong offglide. This type of representation can capture some of the modern-day remnants of historical /r/ codas, as in the diphthongal variants of the \textsc{near} vowel. The offglide may then gradually assimilate to the preceding nucleus, as we see for instance with monophthongal variants of \textsc{near}, such that the only remnants of the offglide gesture are phonological length and phonetic duration. While not all long monophthongs descend from coda liquids, historical coda /r/ vowels, like \textsc{start}, \textsc{nurse} and \textsc{near} exemplify cases in which vowel length descends from a distinct gesture, potentially reinforcing analysis of vowel length in terms of articulatory sequences of vowels.

Some of the theoretical advantages of the two-target model of long monophthongs also set it apart from a third alternative sketched out in Section \ref{theoretical_issues}, namely a model that treats both diphthongs and long monophthongs as inherently dynamic but not compositional, along the lines of \cite{xu2023}. We have not modelled this scenario explicitly, as doing so would represent a major computational challenge and a considerable leap relative to extant models. However, we wish to note some  arguments why this added computational complexity might not provide an improvement over a simpler two-target model. A key difference between a two-target model and a target + slope model is that the vowel offglide is a phonological object in the former case, but not in the latter. In a target model, offglide is a target, i.e. it is corresponds to a phonological object. In a target + slope model, an offglide arises as a phonetic realisation of a gradient slope parameter setting, but it is not a category in itself. Diphthong offglides, however, frequently show some category-like behaviour. As shown by \cite{hsieh2017}, many (but not all) diphthongs can be modelled as a composition of targets that function independently within the same vowel system. Indeed, in our own model, the offglide /i/ target has the same basic parameters as the onglide target of \textsc{fleece} with further gradient variation in the constriction degree, which mirrors such variation arising due to independent phonetic factors. Furthermore, diphthong offglides have a tendency for systemic behaviour. A striking example for this comes from the pattern of changes affecting the \textsc{goose} and \textsc{goat} vowels in English. As noted by \citet[p. 208]{labov1994}, fronting of the \textsc{goose} vowel frequently triggers subsequent fronting of \textsc{goat}. In a two-target model, this can be straightforwardly captured as a generalisation of fronting: fronting of the \textsc{goose} offglide is generalised to the \textsc{goat} offglide. In contrast, it is not so clear how this type of systemic pressure could be captured in a model where vowel dynamics is derived from adjustment to the slope. 

We do acknowledge that our articulatory model we have presented also has some limitations. First, we do not claim to propose optimal parameters for capturing the dynamics of long monophthongs, but instead show that we can reproduce qualitative distinctions that are evidenced in empirical data. Note also that our model assumes a single idealised speaker, but gestural parameters such as TBCD are not necessarily speaker-invariant or language-invariant. The model we propose here allows variation in gestural targets as a mechanism for gradient diphthongisation. We note that such a phenomenon is well captured by dynamical phonological models in which gestural parameters are assigned for an utterance via the evolution of dynamic fields, which correspond to lexical representations \citep{kirov-gafos2007}. A proof-of-concept demonstrating a dynamic neural field model for gradient diphthongisation is presented in \citep{kirkham-strycharczuk2024}. Such a perspective facilitates gradient word-specific phonetic realisations, as well as the potential for dynamic change in phonological representations within and across speakers. We also acknowledge that the key argument for our proposal rests on the assumption that articulatory gestures have intrinsic timing. While a foundational assumption in AP/TD, this is an assumption nonetheless, and it has been called into question \citep{turk2020}.

We would argue that a two-target model of long vowels is promising in that it provides a way of reconciling categorical and gradient aspects of diphthongisation, and it also captures a range of phonetic, phonological and variationist facts. A reanalysis of variation and change affecting degree of diphthongisation in terms of changes to the quality of vowel target is yet to be verified through more systematic modelling. It remains to be seen whether manipulating gestural location and constriction can yield a plausible model of synchronic and diachronic differences in diphthongisation. Another aspect which requires refinement is that of exact timing of the component gestures. A reviewer points out that some varieties of English are distinguished by the timing of the diphthongs, e.g. for some speakers of Australian English \emph{beet} and \emph{boot} are produced with a long onglide, such that the target is reached late. Therefore, a complete model of vowel representation would need to capture the variation not just in the quality of the component gestures, but also in their relative timing. Another compelling question concerns distinguishing between diphthongs and vowel hiatus sequences in articulatory terms. Sequences of adjacent vowels may be heterosyllabic, showing distinct acoustic and phonological properties from diphthongs \citep{chitoran2007}. Diphthongs pattern distinctly from segment sequences, acting as unit, For example they are not reversed in backward talking \citep{cowan1985}. \cite{shaw2021} propose an articulatory framework for distinguishing between complex segments and segment sequences, but their proposal involves in-phase coupling for complex segments, and as such, it is not readily extendable to the diphthong case, where the relevant component gestures are coordinated anti-phase. 

Furthermore, we may ask to what extent the current model extends to models of diphthongs in other languages. The model we have proposed is intended for English, as it is motivated by language-specific phonetic and phonological behaviour. A similar model could be appropriate for long monophthongs in other languages that share phonological and phonetic similarities with diphthongs in the same language. However, there are languages where long monophthongs and diphthongs differ in crucial ways. For instance, Dutch /i/ and /u/ are phonologically long, as shown by their phonotactic behaviour, but they are phonetically short \citep{adank2004dutch}. This differentiates them from /i/ and /u/ in English, as well as from canonical diphthongs in Dutch, for which increased phonetic duration goes hand-in-hand with phonological length. As such, Dutch present a compelling future case study.

 A consequence of our model is that diphthongs are not a distinct phonological category from monophthongs, because diphthongs and long monophthongs share the same phonological structure. In this view, diphthongs can be defined as a sequence of vowels coupled anti-phase that differ in their gestural specifications, whereas the component vowel targets for long monophthongs share the same gestural properties. This is well-supported on the representational level by our phonological modelling. However, identifying individual vowels as diphthongs and monophthongs may prove difficult in practice, because the spatial difference between two component targets can be phonetically very small, corresponding to a small degree of inherent change. There is a question of how much change is needed to be taken as meaningful, and it is not clear that a principled approach is possible, or indeed required in the absence of distinct phonological behaviour that would separate diphthongs from long monophthongs. Accordingly, the terms `monophthong' and `diphthong' become relevant only in the context of phonetic description - they can be used gradiently (`more or less monophthongal / diphthongal'), or to denote typical examples.
 
 A practical consequence is that it is not advisable to assume a distinction between long monophthongs and diphthongs, as is common methodological practice, for instance in acoustic studies of vowel variation. Performing acoustic measurements of vowels frequently involves a degree of dynamic reduction, and it might seem sensible to approach such reduction differently, depending on how much inherent change there is in a vowel. A common approach is taking measurements at a single timepoint for monophthongs and at two timepoints for diphthongs. However, the in-between cases as documented here demonstrate that dynamic differences are likely to emerge within a vowel category. A more principled approach is to use a minimum of two time point measurements for all long vowels, or using a data-driven approach to dynamic reduction (such as fPCA used here or Discrete Cosine Transformation). This may be superfluous in some cases, since a categorical distinction between monophthongs and diphthongs may emerge in some systems, but such a distinction should not be taken for granted. 
 
 \section{Conclusion}
 This study set out to establish whether diphthong vowels can be distinguished from monophthongs by the number of underlying articulatory gestures: two in the case of diphthongs, one in the case of monophthongs. The two-target representation for diphthongs is consistent with the kinematic properties of vowels as documented in this paper, and it is also supported by the results of computational simulation we presented. However, we have argued that the presence of two targets does not in itself entail diphthongisation, but rather, its main consequences are phonological weight and phonetic duration. Therefore, all phonologically long vowels can be modelled as compositional, i.e. having two targets. Within long vowels, degrees of diphthongisation can then emerge from the level of dissimilarity between the two component articulatory targets. The implication of this proposal is that long monophthongs and diphthongs are structurally identical, and therefore they are also predicted to share the same phonological behaviour. Another prediction is that any long vowel could in principle diphthongise, and that diphthongs and long monophthongs can morph into each other fairly freely. These predictions are largely consistent with our current knowledge of diphthongs in British English, but a systematic validation of the proposal would have to take the form of a detailed simulation of variation within the bounds of a particular vowel system showing that varying gestural specific parameters without manipulating the number of gestures can produce realistic patterns of variation. Another compelling avenue for follow-up research is a typological survey of diphthongisation, informed by articulation as well as phonological patterning and historical change, since the interplay between gestural specification, phonological weight and phonetic duration are likely to be language-specific.

 \section*{Acknowledgments}

We thank the participants for contributing to our study. We also thank the \emph{Journal of Phonetics} editor, Taehong Cho, and reviewers for insightful comments on an earlier version of this manuscript.  We are grateful to Ricardo Berm\'{u}dez-Otero, who provided helpful ideas and references, and to Darin Flynn for bringing the case of backward talking to our attention. We are also grateful to the audiences at the Third Angus McIntosh Centre Symposium, the 30th Manchester Phonology Meeting, and the 14th Language Variation and Change Conference for their questions and suggestions. Any remaining errors are our own.  The research was funded by the Arts and Humanities Research Council grant no. AH/S011900/1.

%\bibliography{diphthongisation_biblio, sam_biblio}
%\bibliographystyle{sole}

\newpage
\section*{Appendix A}

\begin{figure}[htbp]
\begin{center}
\includegraphics[scale=1]{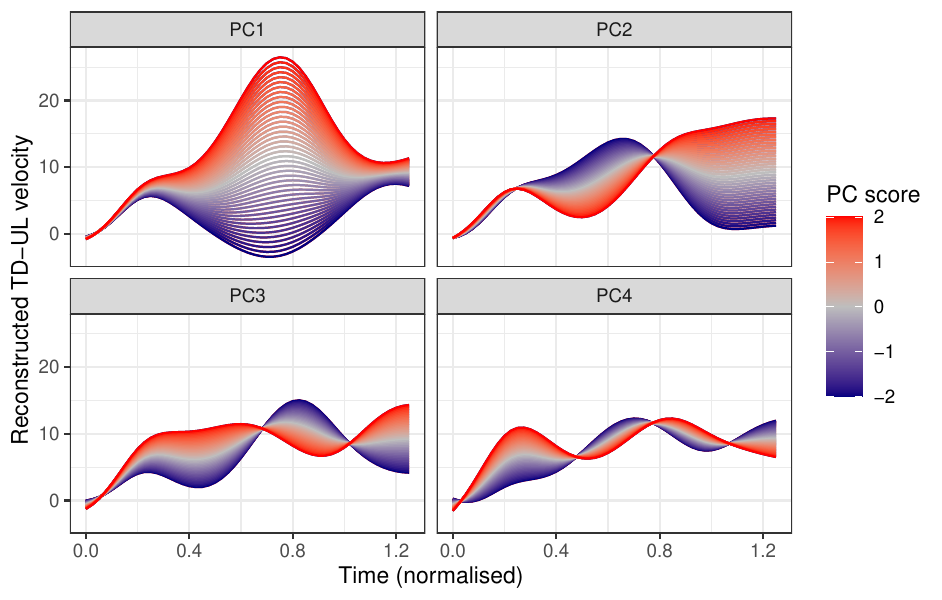}
\caption{Perturbation arising in the TD-UL velocity trajectory as a result of variation in the first four PC scores. The cumulative variance explained with the inclusion of subsequent PCs was: 0.63 (PC1), 0.85 (PC2), 0.97 (PC3), 0.98 (PC4).}
\end{center}
\end{figure}
\newpage

\section*{Appendix B}

\begin{figure}[htbp]
\begin{center}
\includegraphics[scale=0.8]{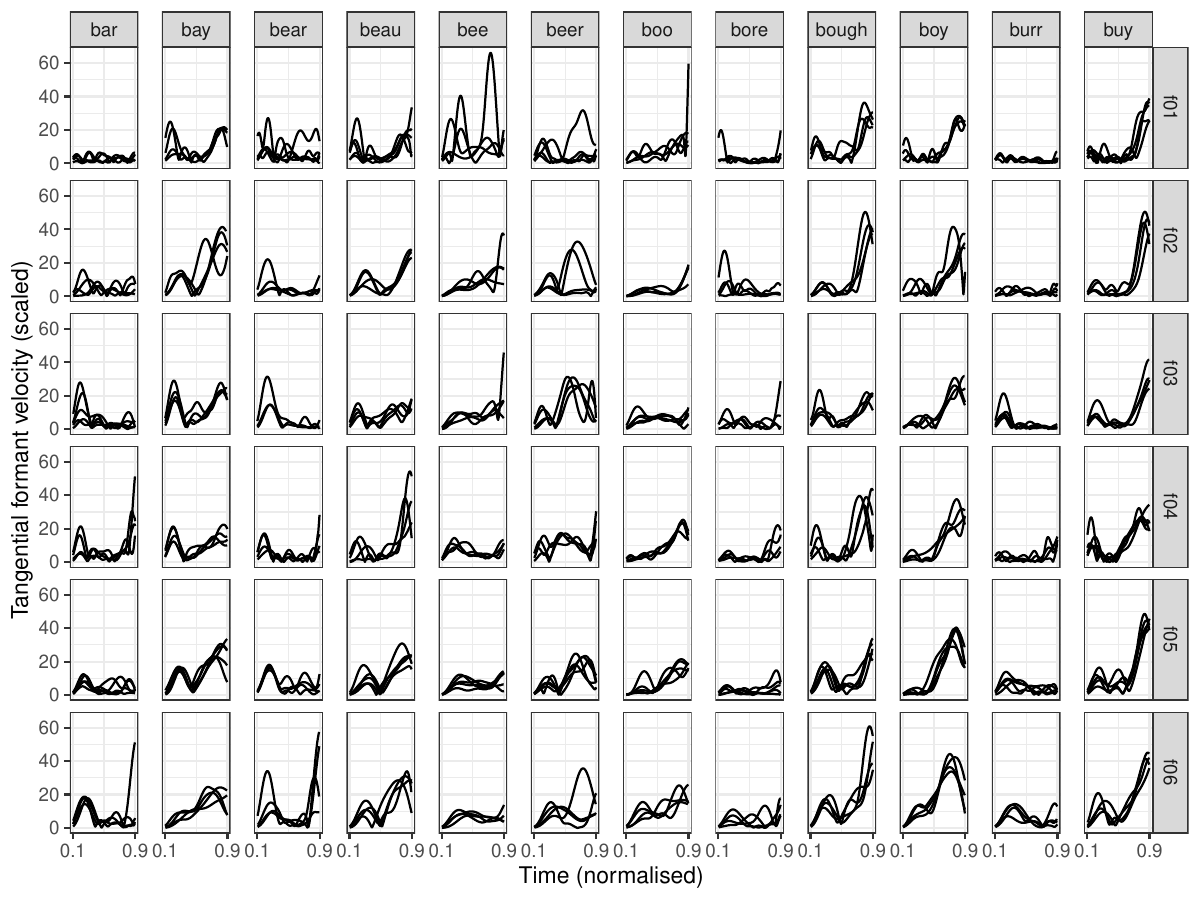}
\caption{Formant velocity profiles for all the individual vowel tokens}
\end{center}
\end{figure}

\newpage
\section*{Appendix C}

\begin{figure}[htbp]
\begin{center}
\includegraphics[scale=1]{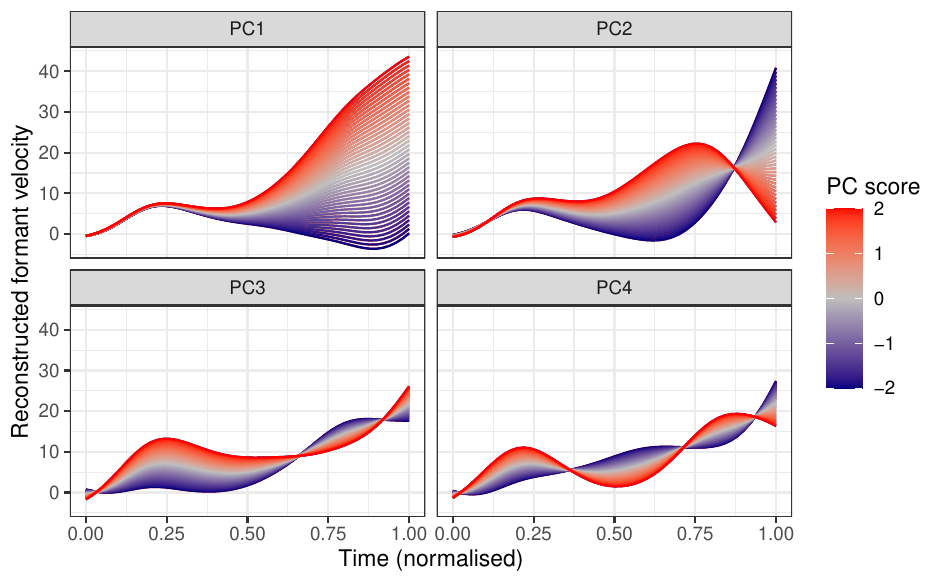}
\caption{Perturbation arising in the formant velocity trajectory as a result of variation in the first four PCs. The cumulative variance explained with the inclusion of subsequent PCs was: 0.63 (PC1), 0.87 (PC2), 0.95 (PC3), 0.99 (PC4).}
\end{center}
\end{figure}

\newpage
\section*{Appendix D}

\begin{figure}[htbp]
\begin{center}
\includegraphics[scale=1]{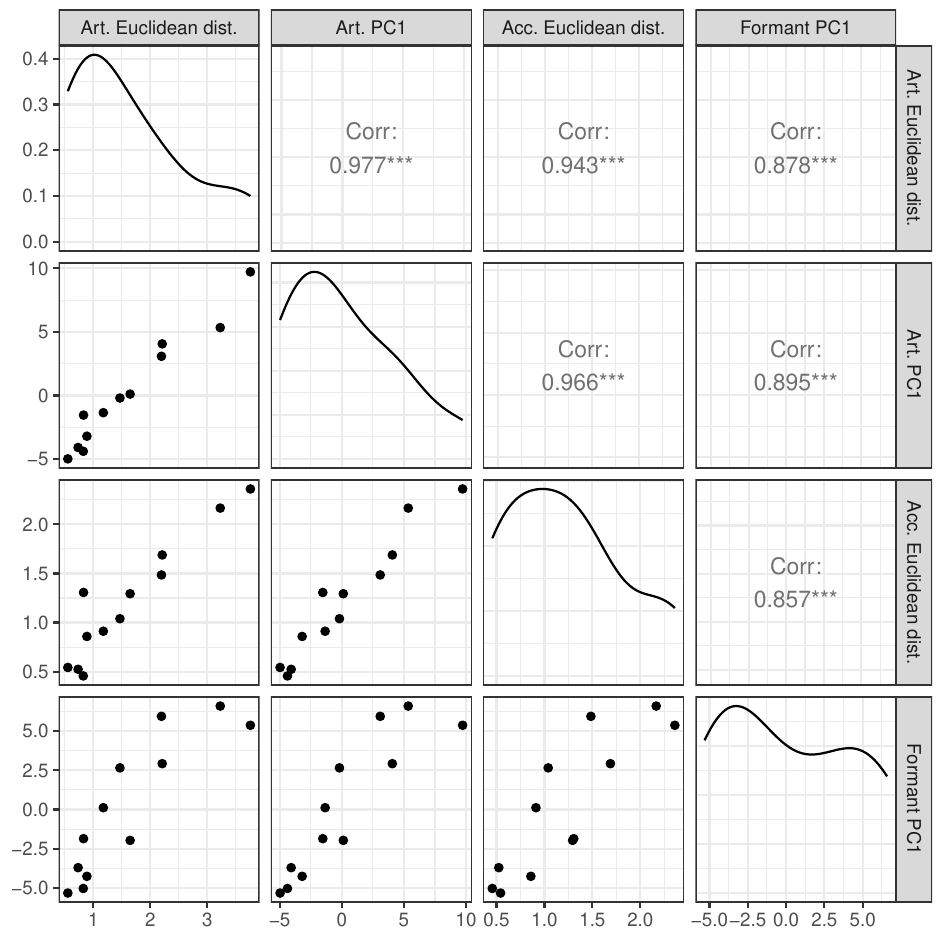}
\caption{Correlations between the by-item means of the four diphtongisation measures.}
\end{center}
\end{figure}

\end{document}